\documentclass[sigconf]{acmart}
\renewcommand\footnotetextcopyrightpermission[1]{} 
\usepackage{amssymb}
\usepackage{algorithm}
\usepackage[noend]{algpseudocode}
\usepackage{amsmath,amsthm}
\usepackage{multirow}
\usepackage{enumitem}
\usepackage{graphicx} 
\usepackage{caption,subcaption}
\usepackage{amsfonts}
\usepackage{ulem}
\let\emptyset\varnothing
\AtBeginDocument{%
  \providecommand\BibTeX{{%
    \normalfont B\kern-0.5em{\scshape i\kern-0.25em b}\kern-0.8em\TeX}}}

\begin{document}
\fancyhead{}


\title{Spectrum-Guided Adversarial Disparity Learning}

\author{Zhe Liu}
\affiliation{\institution{University of New South Wales}}
\email{zhe.liu1@student.unsw.edu.au}

\author{Lina Yao}
\affiliation{\institution{University of New South Wales}}
\email{lina.yao@unsw.edu.au}

\author{Lei Bai}
\affiliation{\institution{University of New South Wales}}
\email{baisanshi@gmail.com}

\author{Xianzhi Wang}
\affiliation{\institution{University of Technology Sydney}}
\email{XIANZHI.WANG@uts.edu.au}

\author{Can Wang}
\affiliation{\institution{Griffith University}}
\email{can.wang@griffithuni.edu.au}



\begin{abstract}
It has been a significant challenge to portray intraclass disparity precisely in the area of activity recognition, as it requires a robust representation of the correlation between subject-specific variation for each activity class.
In this work, we propose a novel end-to-end knowledge directed adversarial learning framework, which portrays the class-conditioned intraclass disparity using two competitive encoding distributions and learns the purified latent codes by denoising learned disparity. Furthermore, the domain knowledge is incorporated in an unsupervised manner to guide the optimization and further boosts the performance.
The experiments on four HAR benchmark datasets demonstrate the robustness and generalization of our proposed methods over a set of state-of-the-art. We further prove the effectiveness of automatic domain knowledge incorporation in performance enhancement. 
\end{abstract}

\ccsdesc[500]{Computing methodologies~Machine learning}
\ccsdesc[300]{Computing methodologies~Neural networks}
\ccsdesc[300]{Computing methodologies~Learning latent representations}
\keywords{Intraclass variability; adversarial autoencoder; activity recognition; generative models}


\maketitle

%

\section{Introduction}
Sensor technologies have inspired a wide range of applications that help people's daily lives in many realms, such as visual recognition \cite{intra_in_face_recoginition}, brain-computer interface \cite{dong2018tri}, and activity recognition \cite{chen2020deep,healthcare}. One of the major components of sensing applications is the establishment of commonly-used and robust systems over diverse scenarios.
Some disparities can be caused by the variety of subjects' temporal conditions and unique physical characters: people have varying habits and body shapes; they may make different movements when performing the same activities, and sensors may perceive activities differently, given the same person performing the same activities. We call such variability within a class \textit{intraclass disparity}.
The intraclass disparity will significantly impair the systems' performance in dealing with new subjects or new environments.

The typical work to address this challenge~\cite{intra_in_face_recoginition} constructs a dictionary or a projection based on the existing subjects to conjecture the possible variations that may occur in less-seen subjects; the models can thus gain robustness through embracing the variation during training. However, such studies primarily rely on the number of known subjects, which we call subject-dependent studies. Therefore, the model may suffer from significant performance degradation when handling new subjects, due to the proprietary characteristics in existing subjects.

In light of generative models' excellent performance on sparse data, recent studies are increasingly applying generative networks to improve models' robustness on new subjects.
Given that generative models usually perform better on sparse data, 
However, most subject-independent studies~\cite{activitysubjectvariance,activitysubjectvariance2,ae_HAR_2,ae_HAR} are still limited in considering the intraclass disparity as meaningless noise and neglect the point that intraclass disparity is related to the subject and the class type. They are still inaccurate in exhibiting the relationship between the subject variation and the class, e.g., subject variation within a class should be conditionally constrained.


Besides, signal data may be segmented imprecisely, and the segments may include gaps and noises. Further, the segments carry nonequivalent amounts of information, which is difficult to measure in the time domain. Since any signal wave in the time domain also presents as a variable in the frequency domain, \textit{Spectrum} represents the corresponding frequency composition of signals. Thus, it is promising to use frequency-domain features (e.g., amplitude and entropy) \cite{yao2017compressive} as the domain knowledge to analyze the segments.
Take amplitude spectrum \cite{signal_Principle} for example. Valid information frequencies offer peak amplitudes than gaps and noises. We can thereby measure the amounts of information and mitigate the impact of imprecise segmentation.
The only issue is the heuristic, hand-engineered, case-specific nature of signal domain analysis, which limits its applicability.


To address the limitations in existing intraclass disparity research, we propose a novel Spectrum-guided Adversarial Autoencoder (SAAE) for Human Activity Recognition (HAR).
HAR is challenging due to its low signal-to-noise ratio. SAAE utilizes two competitive encoding distributions to make intraclass disparity class-conditioned and further eliminates the subject variation through a denoising structure. This way, the purified latent codes will be robust to handle new subjects.
The advantages of SAAE are two-fold. First, SAAE applies a learnable prior distribution, which enables posterior distribution to learn class-related distribution. The posterior distribution learns to generate the intraclass disparity with the class information in prior distribution, while the discriminator and decoder further supervise the prior distribution to learn the pure information by fixing posterior distribution.
Second, SAAE automates spectrum analysis as domain knowledge, leading to enhanced robustness of the data segmentation. The proposed automatic spectrum guided function is capable of dynamically weighing signals and then adjusting the optimization of adversarial training.

\begin{figure*}[t]
  \centering
  \includegraphics[width=0.8\textwidth]{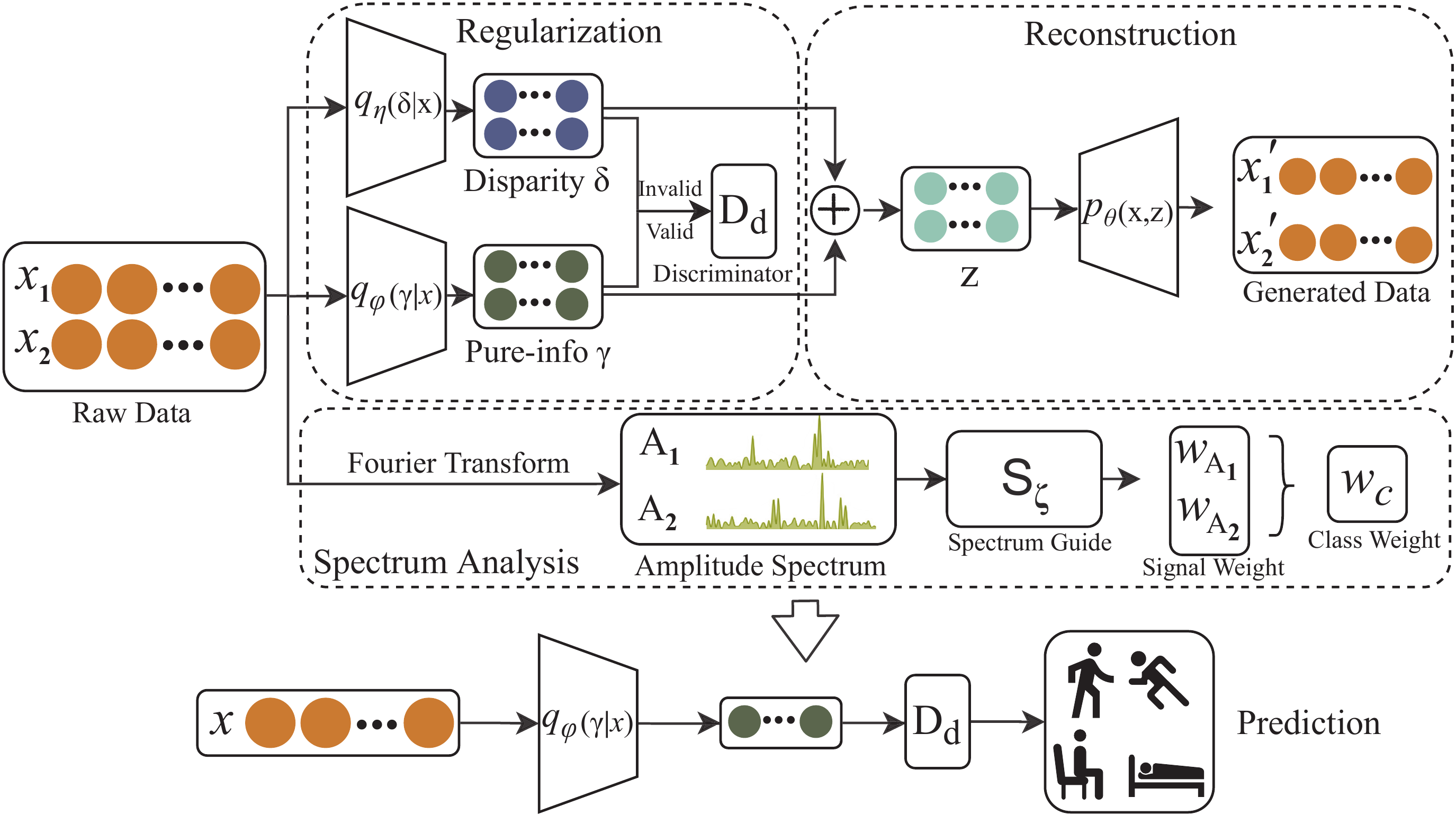}
  \caption{Illustration of SAAE. First, we take two samples of the same class to explain the three phases in optimization: signal weight measurement in spectrum analysis, decomposition in the regularization phase, and denoising in the reconstruction phase.
  \textit{Spectrum analysis} (in the middle) transforms signals and projects the corresponding amplitude spectra into signal weights $W_{A}$ for reconstruction by $S_{\zeta}$. The class weight $W_{c}$ is summarized from all signals under a class to assist regularization.
  \textit{Decomposition} feeds the disparity $\delta$ (labeled `invalid') and pure information $\gamma$ (labeled `valid') to the discriminator to decompose the original data. \textit{Denoising} takes $\delta$ as noises and uses an autoencoder to denoise the disparity components. 
  Then, after optimization, the encoding distribution $q_{\varphi}(\gamma|x)$ could extract information $\gamma$ purified from intraclass disparity $\delta$, which enables discriminator $D$ to precisely predict the activities.
  }
  \label{overviewX}
\end{figure*}

    
We make several contributions in this paper:
\begin{itemize}
    \item We introduce the signal theory and design a principled unsupervised score function to weigh signals dynamically. The score function analyzes signals in the frequency domain to measure the signal information amount and to provide domain knowledge in optimization.
    \item We propose a novel Spectrum-guided Adversarial Autoencoder (SAAE), which fuses automating spectrum analysis and adversarial training in a unified network. SAAE utilizes two competitive encoding distributions supervised by prediction validity to extract and denoise the learned intraclass disparity. We also analyze and prove the effectiveness of intraclass disparity learning through competitive encoding distributions training.
    \item We compare SAAE with both state-of-the-art adversarial autoencoders and human activity-related algorithms in the subject-independent experiments. The superior performance of SAAE on four benchmark datasets demonstrates its effectiveness in solving the intraclass disparity problem on unknown subjects. We further exhibit the effectiveness of intraclass disparity learning through convergence analysis and spectrum analysis.
\end{itemize}

\section{Methodology}



This section introduces the methodology of AAE for the class-conditioned intraclass disparity and the domain knowledge assistance. AAE consists of two phases: intraclass disparity extraction and denoising. AAE leverages the domain knowledge obtained from frequency-domain analysis and an automated spectrum guide function for optimization.

\subsection{Intraclass Disparity Learning}
This section first declares the class conditioned intraclass disparity definition, and then proposes a two-phase AAE which utilizes two competitive encoding distributions to handle the intraclass disparity learning. We discuss the \textit{Reconstruction phase} and the \textit{Regularization phase} of AAE from the generator's perspective and the discriminator's perspective, respectively. We further prove AAE's effectiveness in learning and denoising intraclass disparity.

\subsubsection{Notations for Adversarial Autoencoder}
Following the notations in Kingma et al.'s work~\cite{VAE}, we denote by $p_{data}(x)$ the observed data distribution, $p(x)$ the model distribution of reconstructing data, $q(z|x)$ the encoding distribution, and $p(x|z)$ the decoding distribution. The distribution $q(z|x)$ encodes the original data into latent codes $z$, which denotes the latent representation for data reconstruction. Suppose that the intraclass disparity $\delta$ confuses pure information $\gamma$ in a linear relationship, we define that the latent representation $z$ is composed by
\[
\begin{gathered}
    \gamma \sim q_{\varphi}(\gamma|x),\quad \delta \sim q_{\eta}(\delta|x)\\
    z:=\gamma+\delta
\end{gathered}
\]
where $q_{\varphi}(\gamma|x), q_{\eta}(\delta|x)$ denote the pure information and disparity encoding distributions, respectively.


\subsubsection{Reconstruction Phase}
In this phase, we use the denoising autoencoder to denoise disparity and to reconstruct the signal waveform. We consider the signal waveform reconstruction with the following requirements:
(i) the decoder should be able to reconstruct original data by $z=\gamma + \delta$; (ii) $\gamma$ should be able to learn the representative and invariant waveform information; (iii) $\delta$ should be the confusion factor in waveform only related to class information.

Given the optimal disparity $\delta$, the reconstruction can be demonstrated as a denoising autoencoder from such disparity. The autoencoder fuses the learned disparity $\delta$ with raw data as inputs and then trains autoencoder to recover the original data. Thus, the model will be less sensitive to the disparity components. %
In conventional denoising autoencoders \cite{vincent2008extracting}, the disparity part in original data is vague; thus, they fuse the noise in raw data (e.g., Gaussian noise). Different from the meaningless noise, the proposed disparity encoder $q_{\eta}(\delta|x)$ can extract the disparity components,
which will be proved to be the disparity with class information in the regularization phase by the adversarial training. Therefore, we move the fusion process after encoding. 

The requirement (i)-(iii) can be explained with the following definitions. Let $\gamma$ be pure latent codes and fix the optimal disparity distribution to let $\delta$ be the constant noise. Then, the purified latent code $z=\gamma$ and the confused code $z^{'}=\gamma+\delta$. The condition probability distribution of $z^{'}$ on x is $q_{\varphi}(z^{'}|x)$. Consider an approximate posterior distribution between $x$ and $z$:
\[
q_{\varphi}(z|x)=\int_{z^{'}}c(z|z^{'})q_{\varphi}(z^{'}|x)d z^{'}
\]
where $c(z|z^{'})$ is the observed probability distribution between confused codes and pure codes. Then, given the decoding distribution $p_{\theta}(x,z)=p_{\theta}(x|z)p(z)$, the lower bound of the autoencoder may be formed in the following way \cite{im2017denoising} by Jensen's inequality:
\[
\begin{split}
 log p_{\theta}(x) &= log E_{q_{\varphi}(z^{'}|x)}E_{c(z|z^{'})}[\frac{p_{\theta}(x,z)}{q_{\varphi}(z|z^{'})}]\\
 &\geq E_{q_{\varphi}(z^{'}|x)}E_{c(z|z^{'})}[log\frac{p_{\theta}(x,z)}{q_{\varphi}(z|z^{'})}]
\end{split}›
\]
Therefore, the lower bound of autoencoder can be sorted into minimizing the negative weighted likelihood $L_{rec}$:

\begin{equation}\label{loss_function_for_rec}
    \min_{\varphi,\theta}q_{\varphi}(\gamma|x)[-log\frac{p_{\theta}(x,\gamma)}{c(\gamma|z^{'})}]
\end{equation}

\subsubsection{Regularization Phase} In this phase, we consider the encoding distribution regularization.
The conventional AAE \cite{ae_HAR_2} only applies binary discriminator to regularize generation distribution as a fixed prior distribution by predicting 'real' or 'fake', which fails to portray the class-conditioned intraclass disparity. Therefore, we propose a multi-label classifier and a learnable prior distribution to improve the conventional AAE
by concluding the following requirements:
(i) discriminator should be able to distinguish $\gamma$ and $\delta$ based on the information validity in prediction; (ii) $\gamma$ represents the generalized class information without subject variation; (iii) $\delta$ represents the subject variation conditioned on activity.

As mentioned above, some intraclass disparity exists due to the motion and body shape variance. The similarity distribution regularization will keep this disparity in the representations. Thus, we derive the original binary classifier to a multi-label classifier to distinguish the validity of learned representations. The representations without disparity should be more robust in predicting different subjects, while the disparity representations will impair the performance. Note that, we only consider the case that the dataset contains the class information; otherwise, the prediction validity cannot be measured.
Then, we can explain requirement (ii)-(iii) by denoting the low-validity distribution in regularization as the disparity distribution and high-validity distribution as the pure information in class prediction. The requirements (i) can be modified to let discriminator predict whether the latent codes are `valid' or `invalid' according to the likelihood of correct predictions. 

First, we discuss the Categorical Cross Entropy (CCE) for multi-label classification:
\begin{equation}
CCE(y,\hat{y})=-\sum_{i}^{C}y_{i}log(\hat{y}_{i})=\log(\hat{y}_{c})
\end{equation}
where $y$ denotes a one-hot probability distribution of ground truth; $\hat{y}_{c}$ represents the discriminator probability of true label $c$.

We can find that CCE loss is equivalent to the negative log-likelihood of the true prediction probability. Thus, minimizing the CCE loss equals maximizing the validity (i.e., the likelihood of correct prediction). We utilize the adversarial regularization to optimize requirements (ii)-(iii) and then conclude three requirements in an adversarial format. 

Given data $(X,Y)$, the competitive distribution (i.e., the encoder of $\gamma$) enables the generator distribution (i.e., the encoder of $\delta$) to learn the class conditioned intraclass disparity by optimizing the loss function $L_{reg}$:
\begin{equation}\label{loss_function_for_reg}
    \min_{\eta}\max_{\varphi,d}E_{\gamma\sim q_{\varphi}(\gamma|x)}[\log D_{c}(\gamma)]+E_{\delta\sim q_{\eta}(\delta|x)}[\log (1-D_{c}(\delta))]
\end{equation}
where $d$ denotes the parameters of the multi-label discriminator $D$; $D_{c}$ denotes the probability of correct predictions, respectively.

\begin{proof}
We set up parameters, $\varphi,d,\eta$, before the optimization, and the network has enough capacity to acquire optimal solutions $\varphi^{'},d^{'},\eta^{'}$. When $\varphi$ is fixed, we can take the pure information encoding distribution as the `real' data distribution. According to the Proposition 2 in Goodfellow et. al's work \cite{GAN}, at each iteration, the generator (i.e., the disparity encoding distribution $q_{\eta^{'}}(\delta|x)$) will converge to the `real' data distribution (i.e., the pure information distribution $q_{\varphi}(\gamma|x)$). Hence, the network can converge when $q_{\eta^{'}}(\delta|x)=q_{\varphi}(\gamma|x)$. In other words, the disparity encoder will learn the extracted class information distribution which regularizes the disparity to be conditioned on class. Then, $q_{\eta}(\delta|x)$ can represent a local optimal solution from last iteration, and the optimal $\varphi^{'}$ will be optimized to let discriminator differentiate the validity of disparity and information. Therefore, $q_{\varphi^{'}}(\gamma|x)$ will generate the pure information apart from the previous disparity validated by $D_{d}$. Thus, when the network converges, we can get a local optimal distribution pair $q_{\eta}(\delta|x)$ and $q_{\varphi^{'}}(\gamma|x)$ to decompose latent codes into disparity and pure information, respectively.
\end{proof}

As mentioned in proof, the optimization $D_{d^{'}}$ converges to precisely predicting $q_{\varphi}(\gamma|x)$ and $q_{\eta^{'}}(\delta|x)$, and $q_{\varphi^{'}}(\gamma|x)$ is optimized to be distinct from $q_{\eta^{'}}(\delta|x)$. In other words, $q_{\varphi^{'}}(\gamma|x)$ is invisible to discriminator $D_{d^{'}}$. Thus, only if $q_{\varphi^{'}}(\gamma|x)$ learns the purified data will $D_{d^{'}}$ predict the accurate classification. To further analyze the loss functions of proposed disparity learning, we specify the details of two competitive encoding training and name the two steps of the min-max optimization of $L_{reg}$ as pure loss and disparity loss according to the corresponding encoding distributions:
\begin{equation}
    \begin{gathered}
        L_{pur}:=\max_{\varphi,d}E_{\gamma\sim q_{\varphi}(\gamma|x)}[\log D_{c}(\gamma)]+E_{\delta\sim q_{\eta}(\delta|x)}[\log (1-D_{c}(\delta))]\\
        L_{dis}:=\min_{\eta}E_{\delta\sim q_{\eta}(\delta|x)}[\log (1-D_{c}(\delta))]
    \end{gathered}
\end{equation}
We exhibit curves of $L_{pur},L_{dis}$ to display the convergence of two competitive encoding losses and the training classification performance of using $D_{d^{'}}$ predicting the purified representations $q_{\varphi^{'}}(\delta|x)$ in Section \ref{ablationstudy} \textbf{Convergence Analysis}, which proves intraclass disparity learning and denoising.

\subsection{Principled Spectrum Analysis}
Considering the gaps and noises inside data segmentation, invalid segments will obstruct the models from learning factual patterns. Most deep learning algorithms only analyze data in the time domain, where it is difficult to analyze information components. Therefore, we introduce the domain knowledge of signal theory to leverage the model optimization according to the information amount
from the frequency domain. This section will introduce the principles of spectrum analysis and the fused Spectrum-guided Adversarial Autoencoder, respectively.

\begin{figure}[h]
  \includegraphics[width=\linewidth]{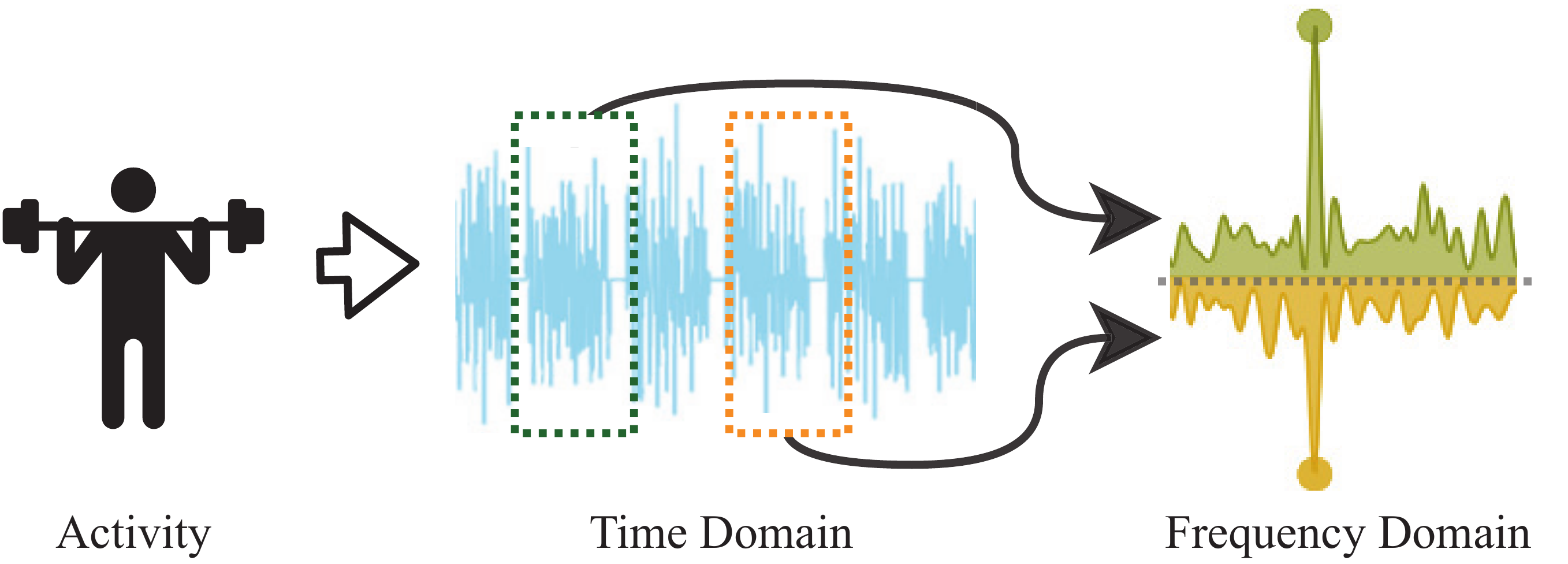}
  \caption{An illustrative example of demonstrating spectrum analysis in the frequency domain. We present two arbitrary data segments while a subject is doing the `weightlifting' activity. It is represented as a typical time-series sequence in the time domain. By introducing the domain-specific knowledge in the theory of signal processing. It can be transformed into a more robust representation by automatic spectrum analysis proposed in this work. The main information of signals (i.e., peak amplitudes in the frequency domain) is indicated with dotted lines.}
  \label{Frequency domain}
\end{figure}

%
\subsubsection{Notations for Spectrum Analysis}
The spectrum analysis focuses on the patterns of signals in the frequency domain, which is transformed from the time domain. Since sensors are only able to record signals in a certain sampling rate, let $x[n]$ be the observed signals in such discrete-time domain. We can obtain the corresponding frequency domain data $X$ and amplitude spectrum $A$ that records the signal strength of each frequency via Fourier transform\cite{DFT}:
\begin{equation}\label{A formula}
    \begin{split}
        X[n] & = \mathcal{F}(x[n]) \\
        & = \sum_{n=0}^{N-1}x[n]\cdot e^{-\frac{i2\pi}{N}kn}\quad k\in [0,N-1]\\
        A[n] & = |X[n]|
    \end{split}
\end{equation}
where $A[n]$ denotes the amplitude of frequency $n$; $N$ denotes the discrete time point number of the sampled signal; $i,e$ represent the imaginary number and the mathematical constant. The amplitude spectrum is perfectly symmetric, so we only analyze the positive frequency part, $k\in [0,N-1]$.


Our goal is to design an amplitude spectrum guide function $S_{\zeta}(A[i])\rightarrow \mathbb{R}$, which can measure the importance of each frequency according to the signal strength. We define the mean score of the amplitude spectrum to represent the signal information amount (i.e., the signal importance) in the numerical format. To ease our illustration, we express the mean score of an arbitrary frequency set $T$ evaluated by spectrum guide function $S_{\zeta}$ as
\[S_{\zeta}(T) = \frac{\sum_{i\in T} S_{\zeta}(i)}{|T|}\]
where $\zeta$ denotes the function parameters; $|T|$ represents the element number of $T$. Therefore, the signal importance can be measured by its amplitude spectrum $S_{\zeta}(A)$.

To automate the spectrum analysis, we analyze amplitude spectrum in terms of the frequency set. According to \cite{signal_Principle}, the signal information is mainly composed of some frequencies with the highest signal strengths. We let $U$ be the frequency set of the highest amplitudes (i.e., information) and $I$ be the frequency set of the lowest amplitudes (i.e., gap and noise):
\begin{equation}\label{UIDefinition}
\begin{gathered}
    \forall i \in U, \forall j \in I, A[i]\gg A[j]\\
    s.t. \quad U \cap I = \emptyset, \quad U,I \subseteq A
\end{gathered}
\end{equation}
where $U,I$ are two subsets of $A$. 

\subsubsection{Spectrum Analysis Principles}
The spectrum analysis principles are based on intra-relationship and inter-relationship in signals, which describe the relationship of frequencies within a signal and between signals, respectively. We illustrate the proposed principles by recalling the phenomenon in Fig. \ref{Frequency domain}:
differing from time-segments, given two arbitrary data segments clipped from a series of noisy continuous signals with gaps, we can easily tell the main composition (i.e., orange and green peak points) and the class of two segments in the frequency domain.
Also, we can easily see that the green segment contains more information, because its curve has more strengths than orange segment's, especially around peak points.
The conventional band-pass filter analysis \cite{signal_Principle} manually selects the frequencies with the highest strengths, i.e., 
the peak points.
To automate the spectrum analysis, we transform the frequency selection to weigh frequencies based on frequency set analysis and propose two principles:

\begin{enumerate}
\item Given an amplitude spectrum $A$, $\forall i\in U, \forall j\in I$, function $S_{\zeta}$ should satisfy \[S_{\zeta}(A[i])-S_{\zeta}(A[j])\propto A[i]-A[j]\]

\item Given the certain frequency's amplitudes from two arbitrary different spectra $A_{a},A_{b}$, $\forall i \in A$, function $S_{\zeta}$ should satisfy \[S_{\zeta}(A_{a}[i])-S_{\zeta}(A_{b}[i])\propto A_{a}[i]-A_{b}[i]\]
\end{enumerate}

The principle (i) describes the traditional band-pass filter theory \cite{signal_Principle}: if the frequency $A[i]$ is the most significant frequency (i.e., the information frequency), $A[i]$ should have much higher amplitudes than others'. Since principle (i) only considers the intra-amplitude difference relationship, we further derive the principle (ii) to complement the inter-relationship.

Beginning with principle (i),
refer to Eq. (\ref{UIDefinition}), we consider that
the \textbf{information set} $U$ and the \textbf{noise set} $I$ (including noise and gaps) meet the significant amplitude difference requirement. To replace the traditional case-specific band selection, we utilize the information set $U$ to represent the frequencies where information may exist. 
Then, we can maximize the below loss function $L_{N}$ to let $S_{\zeta}$ satisfy principle (i):
\begin{equation}
\max_{\zeta} S_{\zeta}(U)-S_{\zeta}(I)
\end{equation}

\begin{proof}
For any two frequencies $i,j$, the score function $S_{\zeta}$ should maximize the weight difference between $i,j$, so we should maximize the loss function that $S_{\zeta}(A[i])-S_{\zeta}(A[j])$. Then, for any two subsets $U,I$, 
\[
\begin{split}
\sum_{i \in U}\sum_{j\in I}S_{\zeta}(A[i])-S_{\zeta}(A[j])&=|U|\cdot|I|\cdot(S_{\zeta}(U)-S_{\zeta}(I))\\
&\propto S_{\zeta}(U)-S_{\zeta}(I)
\end{split}
\]
where $|U|,|I|$ denote the element number subsets. Thus, optimizing $S_{\zeta}(U)-S_{\zeta}(I)$ is equivalent to the raw formula.
\end{proof}

Further, we utilize principle (ii) to distinguish different signal weights. Similar to principle (i), we can easily conjecture the amplitude difference relationship between the particular frequencies of two spectra in the frequency-set format:
\[
\begin{split}
    S_{\zeta}(A_{a})-S_{\zeta}(A_{b})&\propto \frac{1}{N-1}\sum_{i}^{N-1}A_{a}[i]-A_{b}[i]\\
    &=\overline{A_{a}}-\overline{A_{b}}
\end{split}
\]

Then, for arbitrary spectra $\overline{A_{a}}>\overline{A_{b}}$, we can minimize the following loss function $L_{O}$ to meet principle (ii):
\begin{equation}
    \min_{\zeta}|(S_{\zeta}(A_{a}) - S_{\zeta}(A_{b})) - \alpha\cdot(\overline{A_{a}} -  \overline{A_{b}})|
\end{equation}
where $\alpha$ denotes the proportionality constant.

\subsubsection{Spectrum Guide Function}\label{score_function}
To put the proposed principles into practice, we further pose a value range in the function $S_{\zeta}(A)\rightarrow (0,1)$ (0 means extremely insignificant and 1 means extremely significant), and thus the optimal maximum of $S_{\zeta}(U)-S_{\zeta}(I)$ should be 1. Maximizing $L_{N}$ is equivalent to minimising $1-L_{N}$. The empirical proportionality constant holds $\alpha=1$. 

With the above empirical setting, the expected range of $A$ should be $(0,1)$. To match the different analysis perspectives, we propose to rescale the amplitudes by min-max normalization in two folds. Assuming that the spectrum set $\mathcal{A}^{n\times m}:=\{A_{1},A_{2},\dots,A_{n}\}$ contains $n$ spectra with $m$ frequency channels. We normalize each spectrum $A_{i}$ in frequency-level and sample-level to be consistent with two principles (Section 2.2). Denote $A_{i}^{N}$ and $A_{i}^{O}$ to represent the normalized data of intra- and inter-relationships for an arbitrary spectrum $A_{i}$, respectively: 
\begin{equation}\label{normalized A}
    \begin{gathered}
    A^{N}_{i}:=\{\frac{\mathcal{A}_{ij}-\mathcal{A}_{i}^{min}}{\mathcal{A}_{i}^{max}-\mathcal{A}_{i}^{min}}:\mathcal{A}_{i}\in \mathbb{R}^{1\times m},j\in [1,m]\}\\
    A^{O}_{i}:=\{\frac{\mathcal{A}_{ij}-\mathcal{A}_{j}^{min}}{\mathcal{A}_{j}^{max}-\mathcal{A}_{j}^{min}}:\mathcal{A}_{j}\in \mathbb{R}^{n\times 1},j\in [1,m]\}
    \end{gathered}
\end{equation}
where $\mathcal{A}_{i},\mathcal{A}_{j}$ denote a row and column vector of $\mathcal{A}$, respectively; $\mathcal{A}_{i}^{min},\mathcal{A}_{i}^{max},\mathcal{A}_{j}^{min},$ and $\mathcal{A}_{j}^{max}$ denote the minimum and maximum of each row or column.
Then, the score function evaluates the $i$th frequency using both intra and inter normalized features by
\[S_{\zeta}(A[i])=S_{\zeta}(\{A^{N}[i],A^{O}[i]\})\]

To ensure $U$ and $I$ satisfy our definitions, we empirically take the 20\% of frequencies with the highest amplitudes as $U$ and the lowest 50\% as $I$ in $A^{N}$. With the min-max normalization, the mean amplitude difference between $U$ and $I$ will be large enough to distinguish the noise and information.

The optimization of the score function $S_{\zeta}$ can be demonstrated by two steps: 
(i) for any amplitude spectrum $A$, \textcolor{black}{use $A^{N}$ to} update $\zeta$ by minimizing $1-L_{N}$; 
(ii) for any amplitude spectrum pair $(A_{a},A_{b})$, \textcolor{black}{use $(A^{O}_{a},A^{O}_{b})$ to} update $\zeta$ by minimizing $L_{O}$.
We can then unify the steps to optimize $\zeta$ by minimizing a pairwise loss function $L_{S}(A_{a},A_{b})$:
\begin{equation}\label{loss_function_for_s}
\begin{gathered}
    L_{S}(A_{a},A_{b})=S_{\zeta}(I_{a})-S_{\zeta}(U_{a})+S_{\zeta}(I_{b})-S_{\zeta}(U_{b})+2\\
    \quad \quad \quad \quad \quad \quad \quad \quad +|S_{\zeta}(A_{a}) - S_{\zeta}(A_{b}) - (\overline{A_{a}^{O}} -  \overline{A_{b}^{O}})|\\
    s.t.\quad A_{a}:=\{A^{N}_{a},A^{O}_{a}\},A_{b}:=\{A^{N}_{b},A^{O}_{b}\}
\end{gathered}
\end{equation}
where the first line denotes $1-L_{N}$ for $A^{N}_{a},A^{N}_{b}$ and the second line is $L_{O}$ for spectrum pair. Note, we adopt a Fully Connected (FC) layer followed by Sigmoid as $S_{\zeta}$ to match the value range in the definition, and then the optimization will be a traditional regression problem, which can be converged \cite{kingma2014adam}.

\subsubsection{Spectrum-guided Adversarial Autoencoder}

Given a set of signals with labels $(X,Y)$, we calculate the spectrum weighted loss function for an arbitrary signal $(x_{i},y_{i})$ as follows:
\begin{equation}
\begin{gathered}
    \min_{\varphi,\theta}S_{\zeta}(A_{x_{i}})L_{rec};\min_{\eta}\max_{\varphi,d}w_{c}L_{reg}\\
    w_{c}=\frac{1}{|X_{c}|}\sum_{x_{i}\in X_{c}}S_{\zeta}(|\mathcal{F}(x_{i})|)\\
    \quad s.t.\quad X_{c}:= \{x_{i}:y_{i}=c\}
\end{gathered}
\end{equation}
where $A_{x_{i}}$ denotes the amplitude spectrum of $x_{i}$; $w_{c}$ denotes the weight of $x_{i}$'s class; $|X_{c}|$ denotes the number of the samples in class $c$. Due to the activity type and the subject variance, the information amounts carried by signals are varying from each other and thus we apply the mean class spectrum weights of the ground truth to leverage the discriminator training.

We apply two weighing methods in SAAE.
Since the reconstruction phase focuses on a single signal waveform, we leverage different signals by its own spectrum. The regularization phase focuses on the class conditioned disparity, so we leverage signals by overall class information amount distribution. The optimization algorithm is shown in Algorithm \ref{algorithm}. 


\vspace{0.5mm}
\begin{algorithm}[h]
  \caption{Adversarial Autoencoder Training}\label{algorithm}
  \begin{algorithmic}[1]
   \State \textbf{Input} Labeled observations $(X,Y)$
   \State \textbf{Output} $q_{\eta}, q_{\varphi^{'}}, D_{d^{'}}, S_{\zeta^{'}}$
   \State Initialize network parameters $\zeta,\theta,\varphi,\eta,d$
   \While {epoch < max epoch}
   \State Sample a minibatch $\{(x_{1},y_{1}),(x_{2},y_{2}),...,(x_{k},y_{k})\}$
   \For{$i \gets 1$ to $N-1$}                    
        \State $A_{x_{i}}\leftarrow |\mathcal{F}(x_{i})|$
        \State Calculate $A^{N}_{x_{i}},A^{O}_{x_{i}}$
   \EndFor
   \State Sample random spectrum pairs $\{P_{1},P_{2},...,P_{k}\}$
   \State $L_{S}^{'}\leftarrow \frac{1}{k}\sum_{1}^{k}L_{S}(P_{i})$
   \State $\zeta^{'}\leftarrow Adam(L_{S}^{'})$
   \State \textbf{For} $i \leftarrow 1$ to $k$ \textbf{do} Freeze $\zeta^{'}$, $w_{A_{x_{i}}}\leftarrow S_{\zeta}(A_{x_{i}})$
   \State \textbf{For} $c \leftarrow 1$ to $C$ \textbf{do} $W_{c}\leftarrow \{w_{A_{x_{i}}}:y_{i}=c\}$, $w_{c}\leftarrow \overline{W_{c}}$
   \State $w\leftarrow \{w_{c}:c\in [1,C]\}$
   \State $L_{rec}^{'}\leftarrow \frac{1}{k}\sum_{i}^{k}L_{rec}(x_{i},w_{A_{x_{i}}})$
   \State $\varphi^{'},\theta^{'}\leftarrow Adam(L_{rec}^{'})$
   \State $L_{reg}^{'}\leftarrow \frac{1}{k}\sum_{i}^{k}L_{reg}(x_{i},w)$
   \State $\varphi^{'},d^{'},\eta^{'}\leftarrow Adam(L_{reg}^{'})$
  \EndWhile
  \end{algorithmic}
\end{algorithm}

\section{Experiments}
In this section,
we report our comparative experiments and ablation studies in subject-independent settings on four real-world benchmark datasets to evaluate SAAE's robustness and the impact of the spectrum guide function. \textcolor{black}{The source code is publicly available\footnote{\url{https://github.com/leo960912/SAAE}.}}.

\subsection{Datasets}
We evaluate our approach using four public real-world datasets: 1) {MHEALTH}~\cite{MHEALTH}, a sports activity dataset that records signals of motion and inertial measurement unit (IMU) sensors for 12 different sports activities \textcolor{black}{of 10 volunteers}; 2) {PAMAP2}~\cite{pamap2}, a daily activity dataset related to 18 daily activities collected from IMUs deployed at different areas of body \textcolor{black}{on 9 subjects}; 3) {UCIDSADS}~\cite{ucidsads}, concerning \textcolor{black}{8 subjects'} 19 daily and sports activities performed with speed and amplitude variations recorded by motion sensors and IMUs; and 4) {OPPORTUNITY}~\cite{oppo}, concerning \textcolor{black}{4 subjects'} 17 hand activities recorded by various body-worn, object-based, and ambient
sensors.
Considering not all subjects performed all activities in {PAMAP2}, we exclude six activities (watching TV, computer work, car driving, folding laundry, house cleaning, and playing soccer) and one subject who conducts very few activities from our experiments.

\subsection{Experiment Setting}
We execute Leave-One-Subject-Out experiments (with the test subject being excluded from training) to evaluate algorithms' robustness on varying subjects, where we set the time windows as 20 with 50\% overlapping to pre-process the time-sequence data. \textcolor{black}{The spectrum data calculated by Fourier Transform are pre-calculated, so the time complexity is O(1).}
%
We compare our algorithm, {\em SAAE} with five state-of-the-art HAR models and three AAE models,
and show a further study on {\em iAAE}, a version of the proposed model without the knowledge (spectrum) guide function.

$\bullet$ \textit{MC-CNN}~\cite{mccnn}: a state-of-the-art Convolutional Neural Network (CNN) that captures temporal correlations along the time axis

$\bullet$ \textit{Bi-LSTM-S}~\cite{2016ijcai}: a bi-directional Long Short-term Memory (LSTM) that captures both forward and backward time information

$\bullet$ \textit{ConvLSTM}~\cite{convlstm}: a hybrid model that combines both CNN and LSTM to capture both the spatial and temporal correlations.

$\bullet$ \textit{En-LSTM}~\cite{ensemblelstm}: an LSTM-based method that combines multiple individual LSTM learners with epoch-wise bagging

$\bullet$ \textit{AttConvLSTM}~\cite{attention-convlstm}: a hybrid model that applies attention layer to learn weighted information

$\bullet$ \textit{AAE}~\cite{adversarialautoencoder}: a benchmark of adversarial autoencoder with FC layers

$\bullet$ \textit{ConvAAE}~\cite{ae_HAR_2}: a state-of-the-art adversarial autoencoder algorithm to learn the long-time information for gesture recognition

$\bullet$ \textit{DAAE}~\cite{creswell2018denoising}: a convolutional adversarial autoencoder enhanced with denoising operation to extract the pure information from low signal-to-noise ratio data

We take MC-CNN as the base model for the encoder, three deconvolutional blocks, named the deconvolutional layer, the rectified linear unit (ReLU), and the batch normalization layer for the decoder, an FC layer, and a softmax function for the discriminator. The network architecture details are shown in the supplementary Section \ref{Architecture}. 
Our model applies the same encoder-decoder structure as in DAAE but differs in using the adversarial structure and the spectrum guide function.
%
To show the robustness of methods, we use the same hyperparameter settings over all the subjects in the same datasets. For SAAE, we set the learning rate of the spectrum guide function to 1e-4 for 4 datasets, the learning rates of the adversarial network to 5e-5 for PAMAP2 and 2e-4 for other datasets. 

\subsection{Overall Comparison}\label{Spectrum_analysis_effectiveness}
\begin{table*}[!htb]
\centering
\caption{Overall Comparison of SAAE over Four Benchmark Datasets}
\label{mainexperiment}
\footnotesize
\resizebox{0.95\textwidth}{!}{%
\begin{tabular}{c|c|ccccc}
\hline
\multirow{8}*{MHEALTH} & Metrics  & MC-CNN & Bi-LSTM & ConvLSTM & En-LSTM & AttConvLSTM \\ 
\cline{2-7}
 ~ & Acc  & 0.927(0.031) & 0.899(0.044) & 0.913(0.038) & 0.860(0.089) & 0.921(0.042) \\ 
~ & Pre & 0.935(0.039) & 0.871(0.062) & 0.893(0.051) & 0.848(0.085) & 0.899(0.052) \\
~ & F1 & 0.922(0.037) & 0.879(0.054) & 0.899(0.047) & 0.846(0.088) & 0.908(0.050) \\
\cline{2-7}
~ & Metrics & AAE & ConvAAE & DAAE & iAAE & SAAE\\
\cline{2-7}
~ & Acc & 0.907(0.054) & 0.826(0.061) & 0.947(0.049) & 0.955(0.028) &\textbf{ 0.958(0.028)}\\
~ & Pre & 0.905(0.067) & 0.804(0.078) & 0.941(0.063) & 0.960(0.026) &\textbf{ 0.963(0.026)}\\
~ & F1 & 0.897(0.063) & 0.809(0.072) & 0.941(0.060) & 0.957(0.028) & \textbf{0.960(0.030)}\\
\hline
\multirow{8}*{PAMAP2} & Metrics  & MC-CNN & Bi-LSTM & ConvLSTM & En-LSTM & AttConvLSTM \\ 
\cline{2-7}
~ & Acc &  0.803(0.133) & 0.715(0.200) & 0.757(0.158) & 0.734(0.157) & 0.741(0.146) \\ 
~ & Pre  & 0.806(0.134) & 0.711(0.241) & 0.730(0.202) & 0.739(0.196) & 0.739(0.120) \\
~ & F1  & 0.781(0.154) & 0.687(0.230) & 0.724(0.192) & 0.720(0.180) & 0.718(0.150) \\
\cline{2-7}
~ & Metrics & AAE & ConvAAE & DAAE & iAAE & SAAE\\
\cline{2-7}
~ & Acc &  0.727(0.194) & 0.713(0.147) & 0.774(0.180) & 0.837(0.112) &\textbf{ 0.840(0.109})\\
~ & Pre & 0.771(0.195) & 0.730(0.155) & 0.805(0.153) & \textbf{0.855(0.096)} & \textbf{0.855(0.101)}\\
~ & F1 & 0.746(0.215) & 0.693(0.168) & 0.764(0.197) & 0.831(0.124) & \textbf{0.836(0.122)}\\
\cline{2-7}
\hline
\multirow{8}*{UCIDSADS} & Metrics  & MC-CNN & Bi-LSTM & ConvLSTM & En-LSTM & AttConvLSTM \\ 
\cline{2-7}
~ & Acc  & 0.879(0.067) & 0.889(0.049) & 0.897(0.046) & 0.831(0.043) & 0.887(0.048)\\ 
~ & Pre  & 0.872(0.095) & 0.897(0.066) & 0.896(0.063) & 0.841(0.065) & 0.882(0.072) \\
~ & F1  & 0.855(0.087) & 0.877(0.061) & 0.884(0.059) & 0.811(0.055) & 0.868(0.064) \\
\cline{2-7}
~  & Metrics & AAE & ConvAAE & DAAE & iAAE & SAAE\\
\cline{2-7}
~ & Acc & 0.846(0.045) & 0.815(0.033) & 0.889(0.047) & 0.918(0.044) & \textbf{0.929(0.040)}\\
~ & Pre & 0.857(0.055) & 0.790(0.047) & 0.907(0.041) & 0.926(0.045) &\textbf{ 0.935(0.048)}\\
~ & F1 & 0.824(0.056) & 0.785(0.040) & 0.875(0.056) & 0.906(0.052) & \textbf{0.919(0.047)}\\
\hline
\multirow{8}*{OPPORTUNITY} & Metrics  & MC-CNN & Bi-LSTM & ConvLSTM & En-LSTM & AttConvLSTM \\ 
\cline{2-7}
& Acc & 0.635(0.050) & 0.575(0.096) & 0.537(0.088) & 0.531(0.107) & 0.566(0.092) \\ 
~ & Pre & 0.637(0.024) & 0.599(0.101) & 0.482(0.124) & 0.534(0.094) & 0.587(0.093)  \\
~ & F1 & 0.613(0.042) & 0.549(0.098) & 0.464(0.116) & 0.485(0.113) & 0.530(0.107) \\
\cline{2-7}
~  & Metrics & AAE & ConvAAE & DAAE & iAAE & SAAE\\
\cline{2-7}
~ & Acc & 0.624(0.061) & 0.609(0.078) & 0.625(0.069) & 0.664(0.056) &\textbf{ 0.680(0.049)}\\
~ & Pre & 0.663(0.025) & 0.661(0.062) & 0.618(0.076) & 0.698(0.053) & \textbf{0.713(0.048)}\\
~ & F1 & 0.603(0.064) & 0.597(0.077) & 0.598(0.075) & 0.655(0.055) & \textbf{0.674(0.046)}\\
\hline
\end{tabular}
}
\label{datasets}
\end{table*}
The results (Table \ref{mainexperiment}) show our models outperforming all other algorithms on the four datasets and having the smallest standard deviations, demonstrating the effectiveness of SAAE and its variant, iAAE (without Spectrum knowledge) on four benchmark datasets. We can easily observe that the proposed models, iAAE and SAAE, consistently outperform all other baselines and hold the smallest standard deviation. In terms of F1-score, SAAE improves about 2\%, 3\%, 4\%, and 6\% in four datasets, It also indicates the robustness in handling intraclass variance of new subjects. 


\begin{figure*}[htb]
    \centering 
\begin{subfigure}{0.24\textwidth}
  \includegraphics[width=\textwidth]{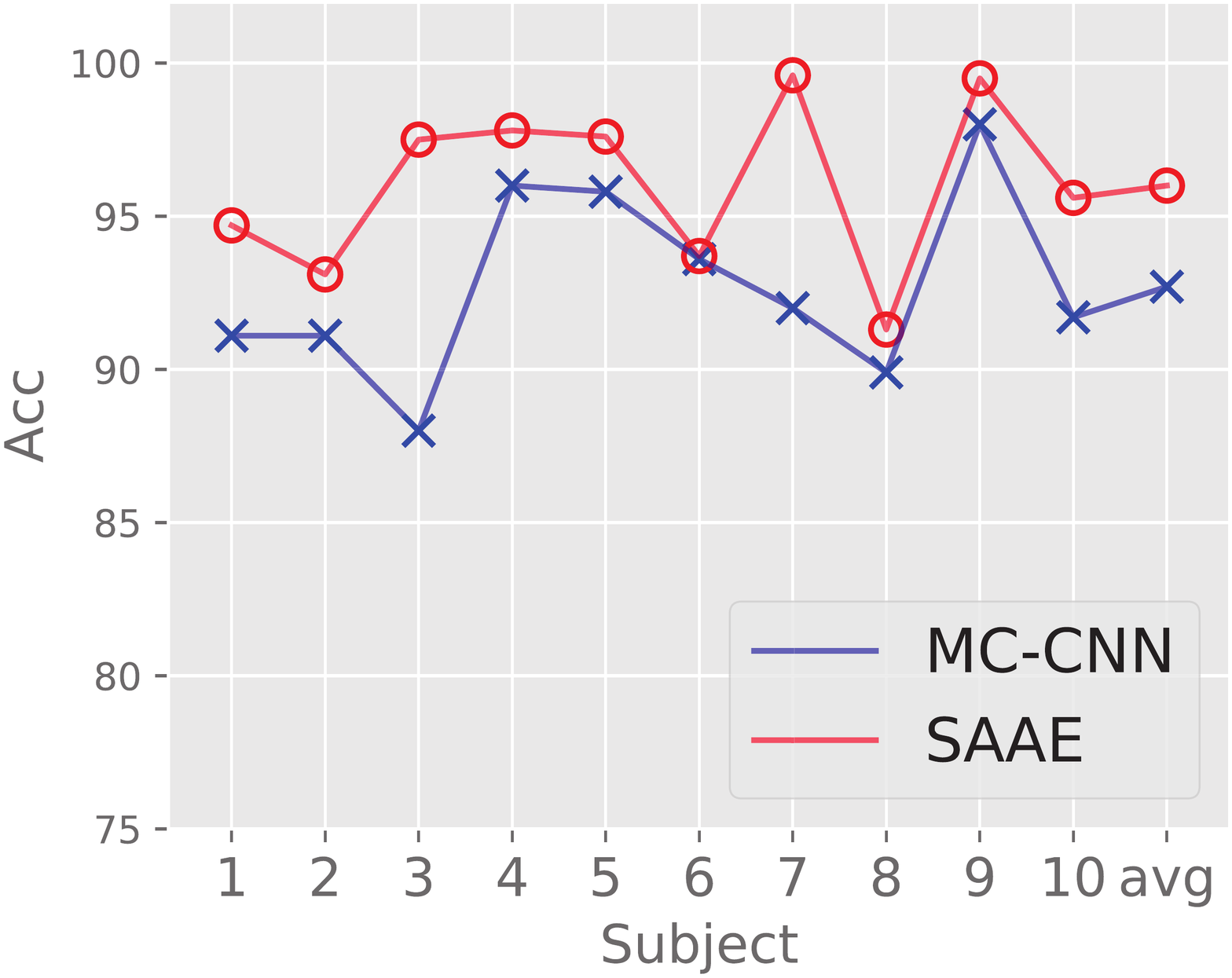}
    \centering
  \caption{ Detailed Accuracy}
\end{subfigure}\hfil 
\hspace{-1mm}
\begin{subfigure}{0.24\textwidth}
  \includegraphics[width=\textwidth]{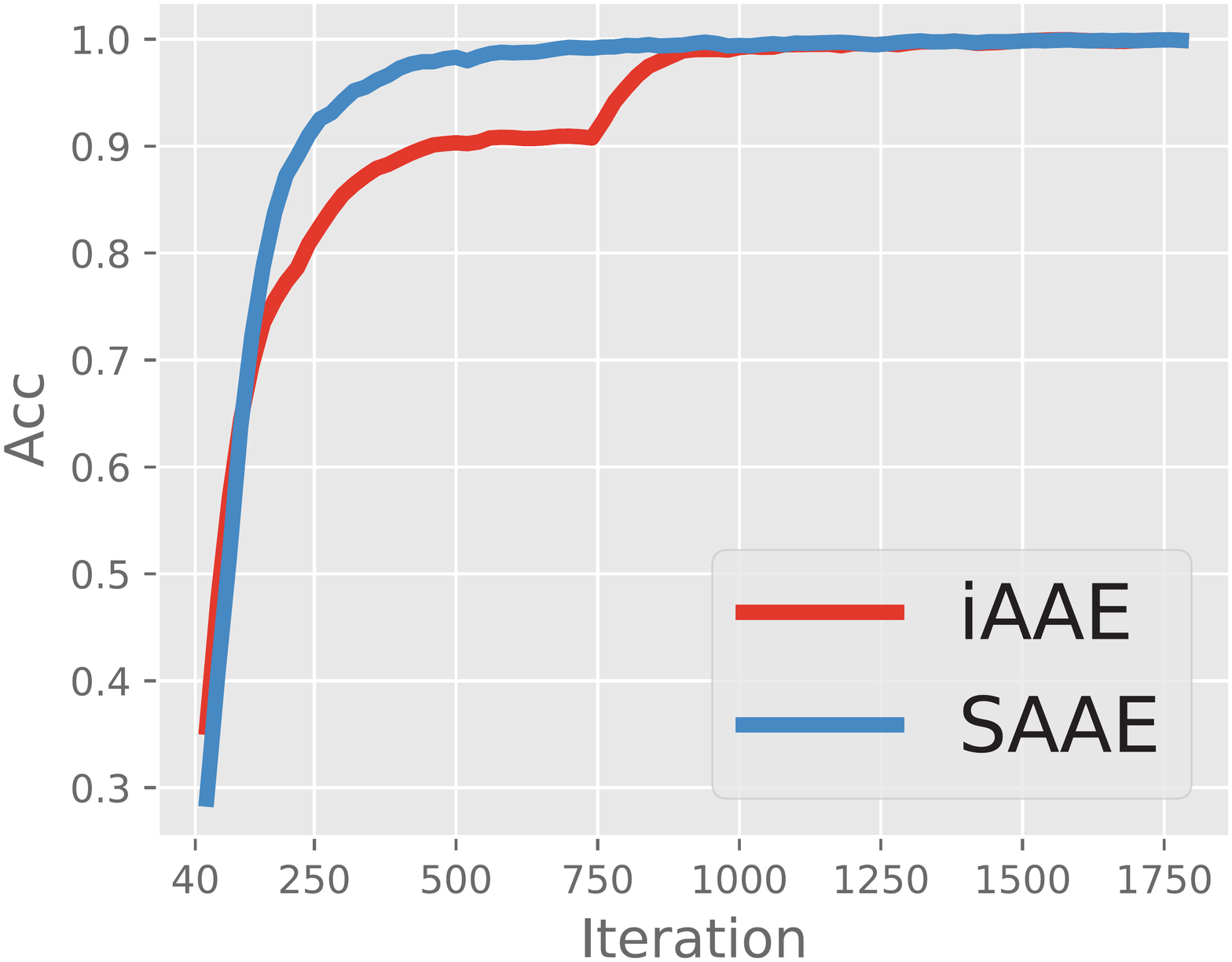}
    \centering
  \caption{Dis Acc Comparison}
\end{subfigure}\hfil 
\hspace{-1mm}
\begin{subfigure}{0.24\textwidth}
  \includegraphics[width=\textwidth]{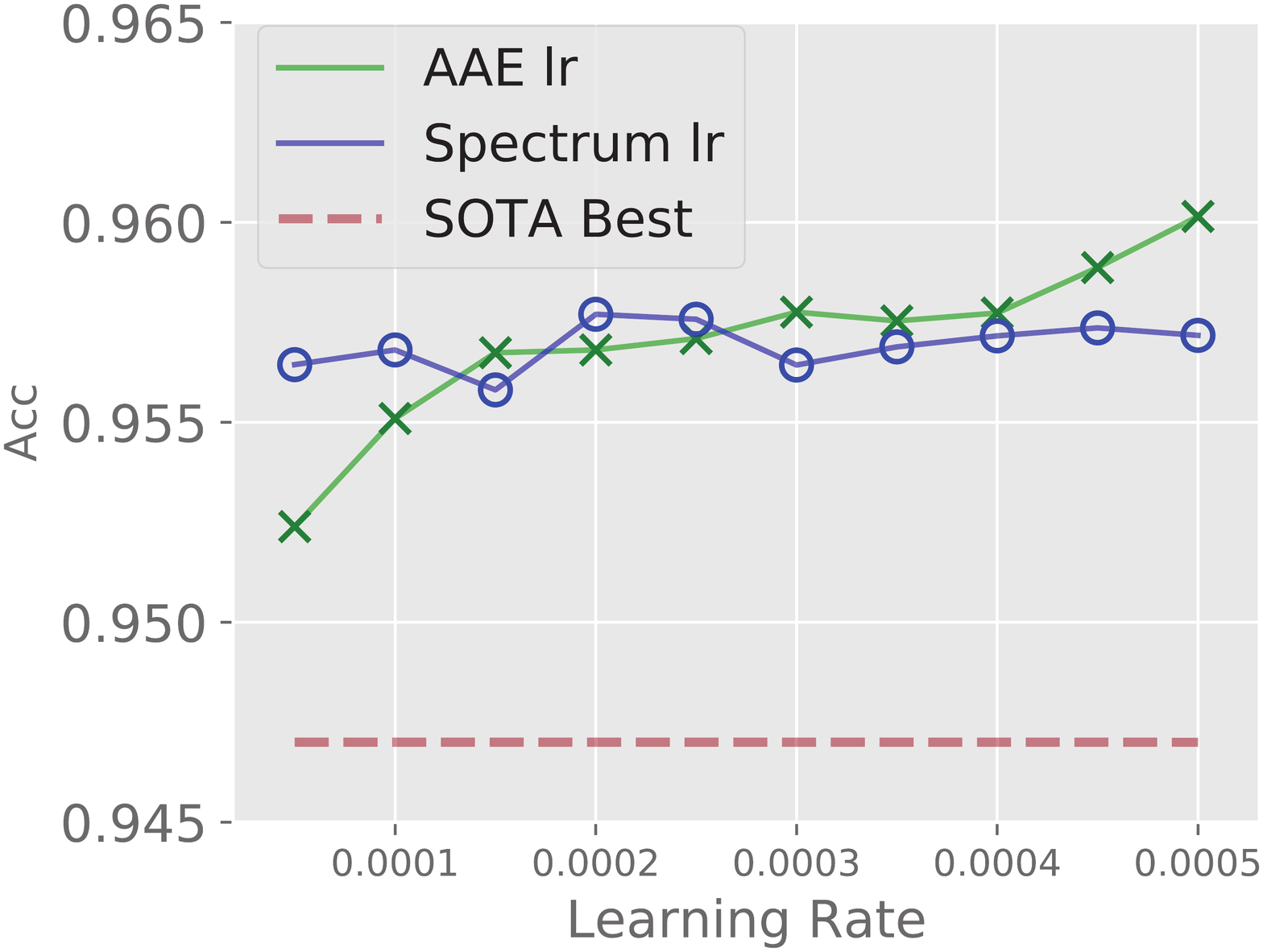}
    \centering
  \caption{ Hyper-parameter}
\end{subfigure}\hfil 
\hspace{-1mm}
\begin{subfigure}{0.24\textwidth}
  \centering
  \includegraphics[width=\textwidth]{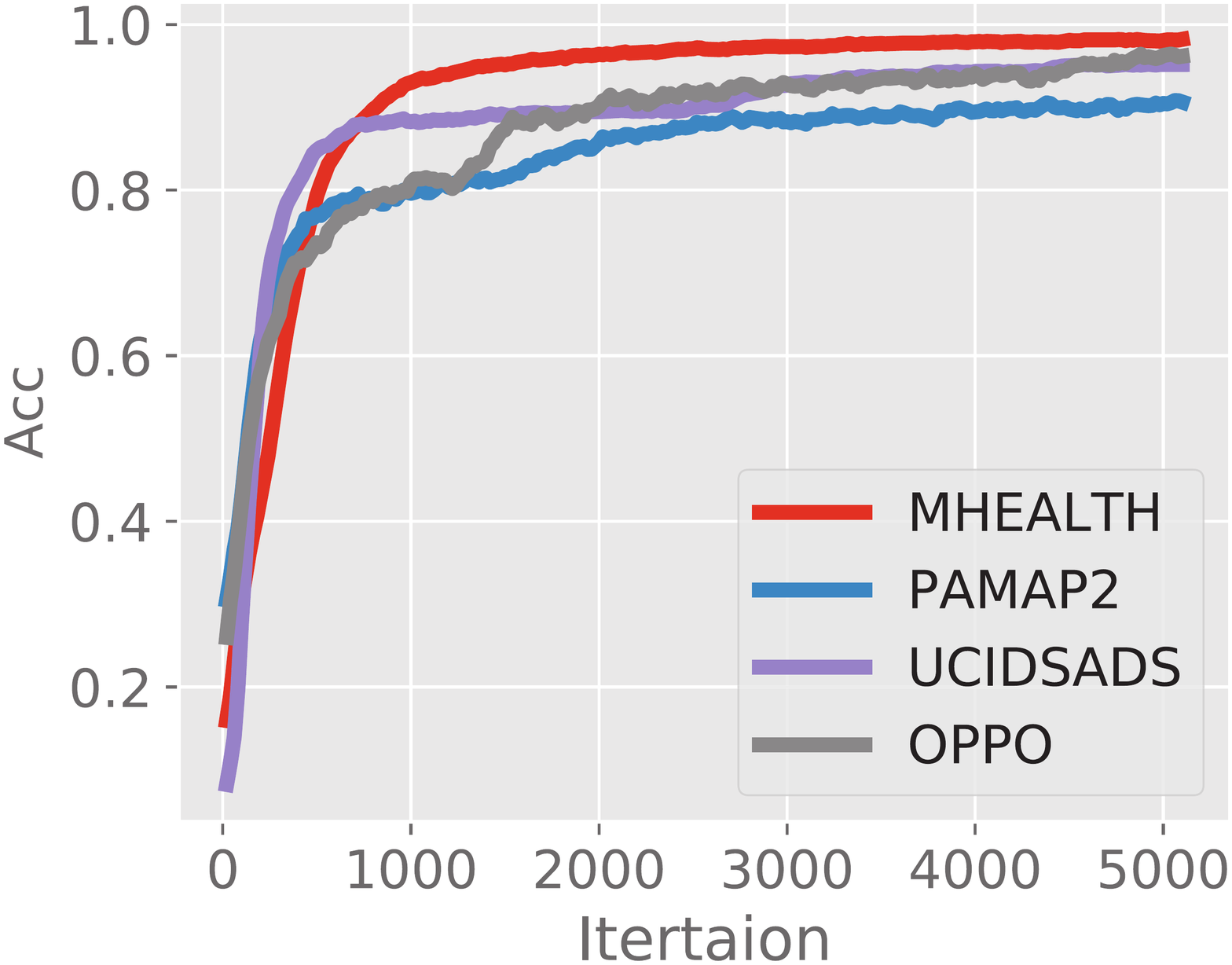}
  \caption{  Discriminator Acc}
\end{subfigure}\hfil 
\hspace{-1mm}
\medskip
\begin{subfigure}{0.24\textwidth}
  \centering
  \includegraphics[width=\textwidth]{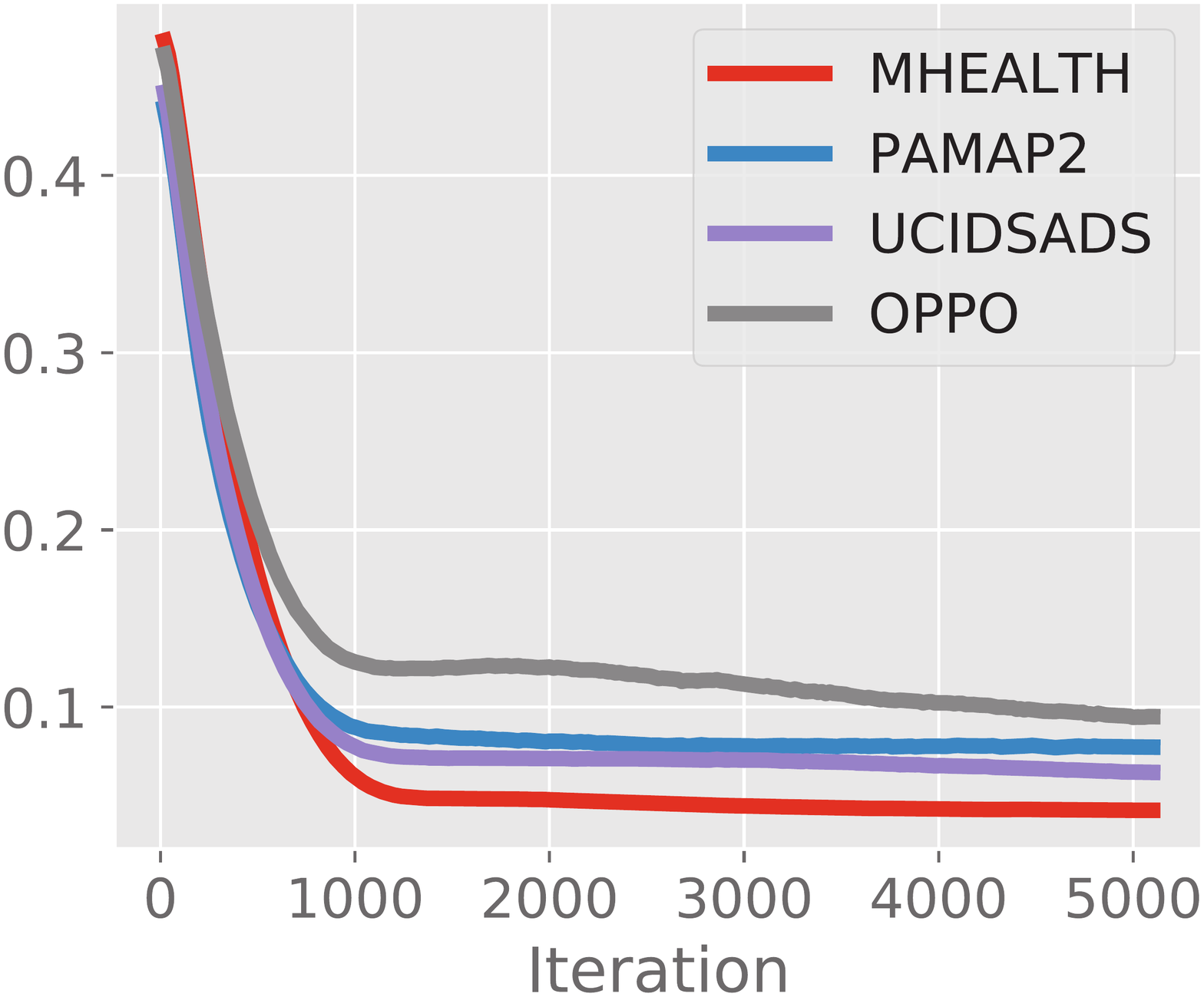}
  \caption{  Loss $L_{rec}$}
\end{subfigure}\hfil 
\hspace{-1mm}
\begin{subfigure}{0.24\textwidth}
  \centering
  \includegraphics[width=\textwidth]{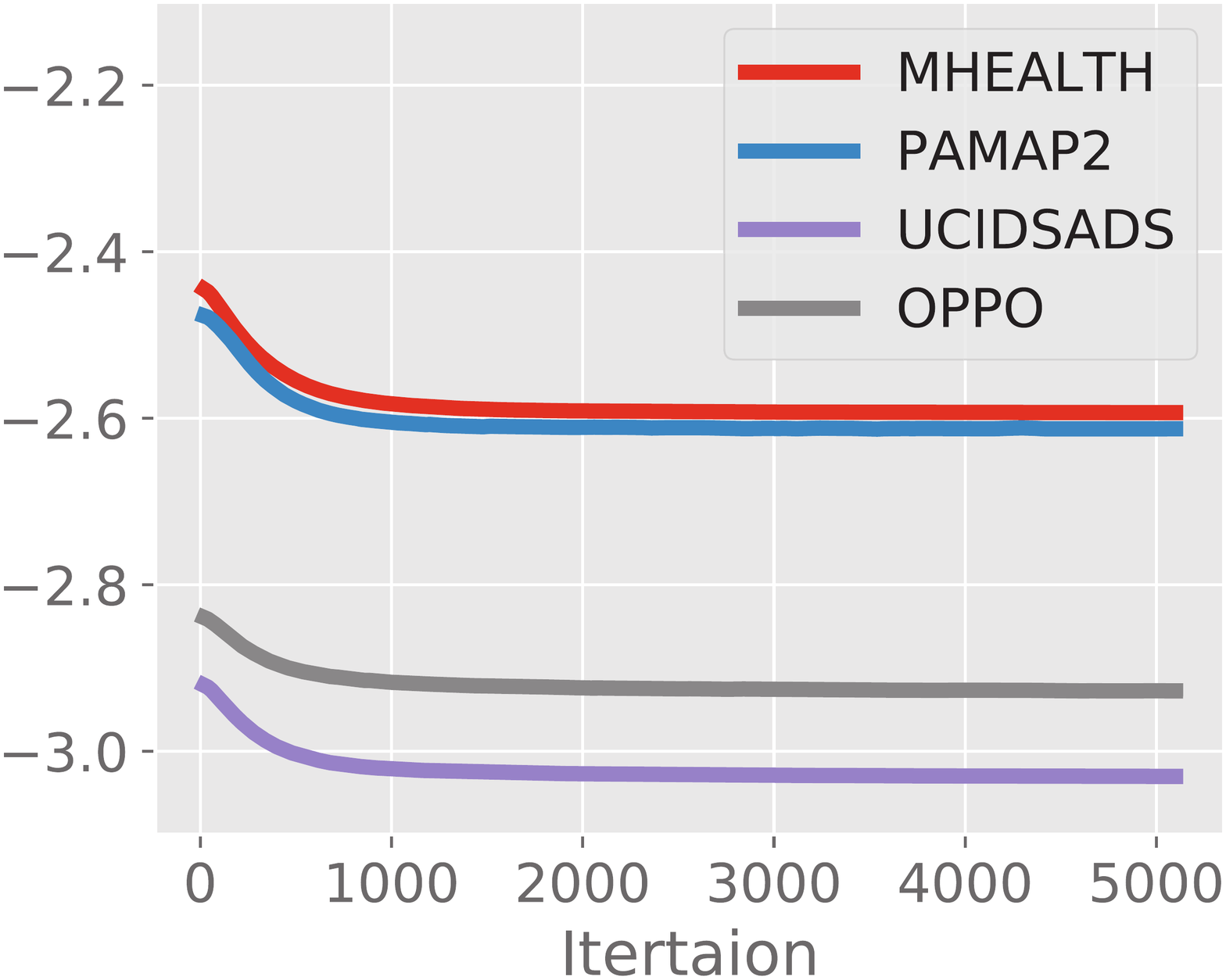}
  \caption{ Loss $L_{dis}$ of $L_{reg}$}
\end{subfigure}
\hspace{-1mm}
\begin{subfigure}{0.24\textwidth}
  \centering
  \includegraphics[width=\textwidth]{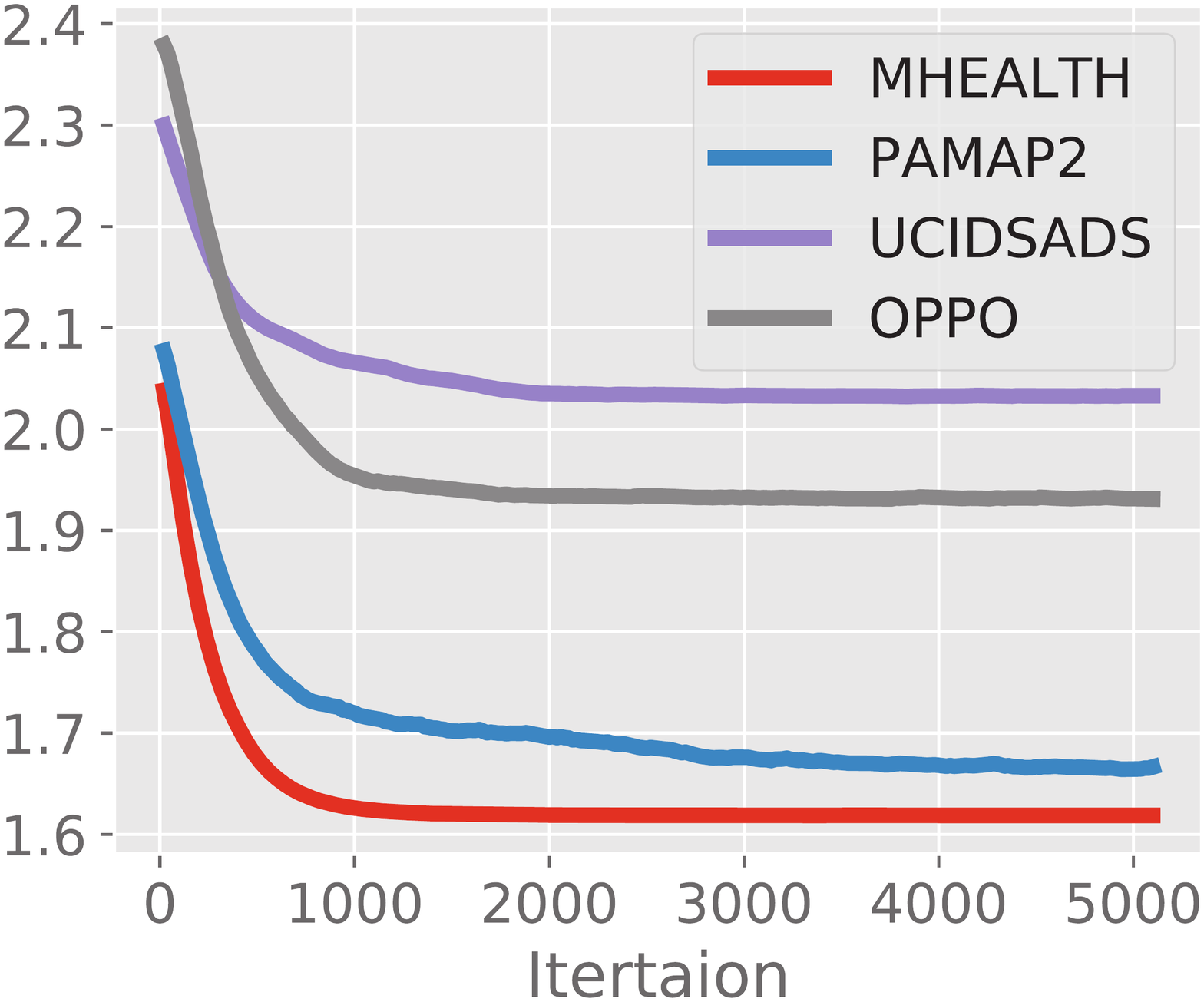}
  \caption{ Loss $L_{pur}$ of $L_{reg}$}
\end{subfigure}
\hspace{-1mm}
\hspace{-1mm}
\begin{subfigure}{0.24\textwidth}
  \centering
  \includegraphics[width=\textwidth]{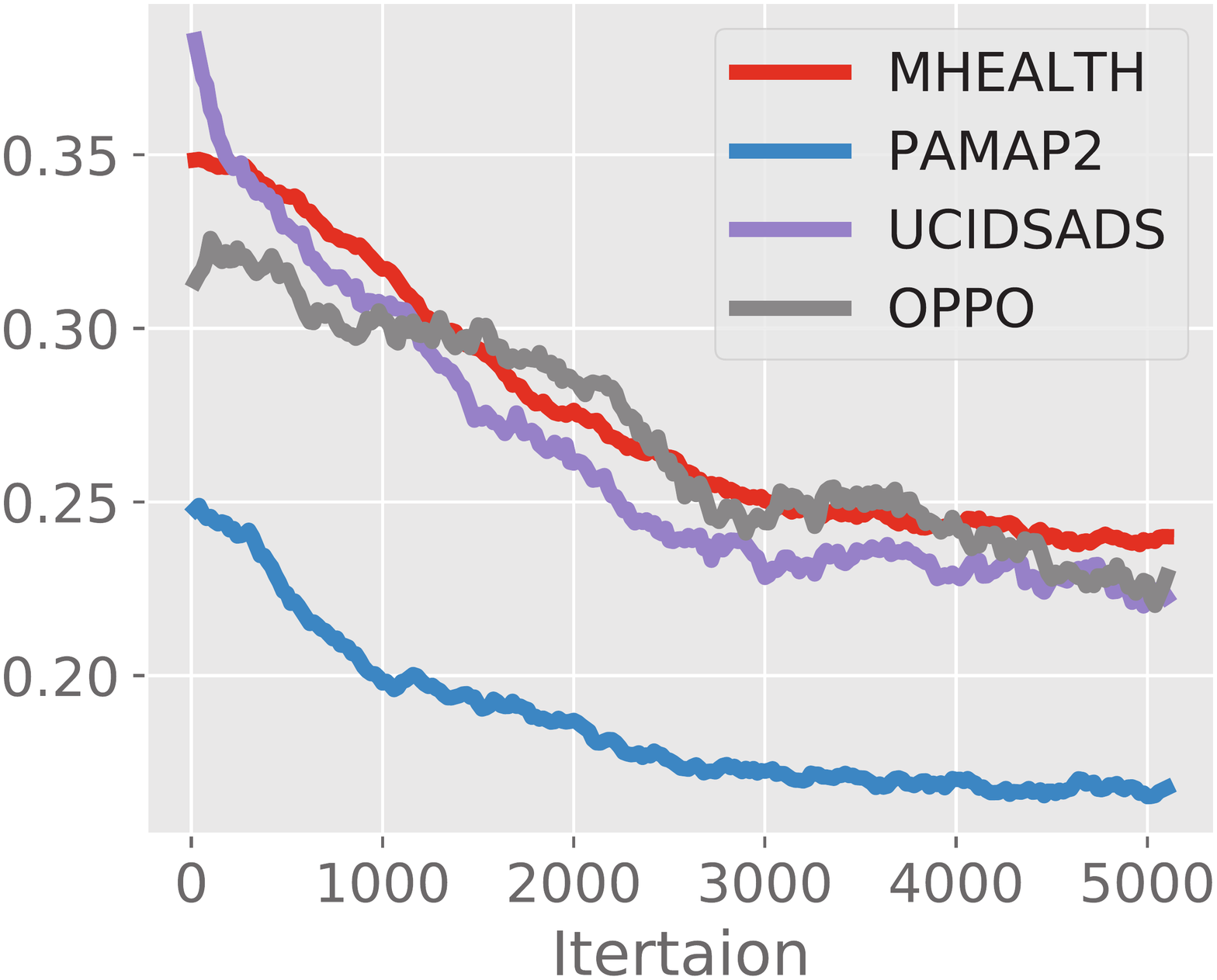}
  \caption{ Spectrum Loss $L_{S}$}
\end{subfigure}
\caption{Experiment Analysis.}
\label{MHEALTH}
\end{figure*}

\subsection{Further Analysis}\label{ablationstudy}
\vspace{.5mm}\noindent\textbf{Effectiveness of Adversarial Training.}
\textcolor{black}{From the mean performance of four datasets,} SAAE achieves a 4.3\% improvement in accuracy, 4.8\% in precision, and 5.2\% in F1 Score comparing with a set of encoder-decoder based methods. It demonstrates the advantage of our proposed competitive encoding distribution learning against conventional adversarial training approaches.

We further show comparisons on MHEALTH at the individual level (Fig. \ref{MHEALTH} (a)). For instance, MC-CNN can achieve 97\% accuracy on subject 9 while only 87\% in subject 3, revealing that MC-CNN may perform badly while dealing with the new subject whose samples are deviating a bit more from the common distribution of training subjects. In contrast, SAAE steadily improves the performance on most subjects, especially subject 3 and subject 7, which proves its robustness and reliability.  

\vspace{.5mm}\noindent\textbf{Effectiveness of Spectrum Analysis.}
Comparing SAAE and iAAE, we find that the spectrum guide function can further enhance the performance by around 1\% in each subject,
validating the effectiveness of frequency domain analysis in model learning. We also take subject 1 in MHEALTH as an instance to illustrate the spectrum analysis's effectiveness in optimization. Our convergence comparison of discriminator $D$ (Fig. \ref{MHEALTH} (b)) shows spectrum domain knowledge can better exclude the disparity during optimization.

\vspace{.5mm}\noindent\textbf{Hyper-parameter Analysis.}
We change the learning rates of different components to explore the optimal hyper-parameters. We set learning rate as 1e-4 for spectrum guide function and 2e-4 for AAE as default, and plots the mean accuracy of one-out subject-independent experiments when learning rates range from [5e-5,5e-4] in Fig. \ref{MHEALTH} (c). We can observe that SAAE is stable over different parameters and constantly outperforms the best state-of-the-art. The spectrum learning rate merely influences the model performance while AAE's learning rate will slightly affects the results. With larger learning rate of AAE, the results gradually become better, which means SAAE could act better than our provided result. 

\vspace{.5mm}\noindent\textbf{Convergence Analysis.}
Fig. \ref{MHEALTH} (d) - (h) plot the averaged loss over four datasets. We can observe that $L_{rec},L_{dis},L_{pur}$ could quickly converge around 1000 iterations and then remain stable, where $L_{dis}$ and $L_{pur}$ represent the min-max competition loss functions of intraclass disparity and purified information, respectively. The convergence of $L_{pur}$ and $L_{dis}$ reveals the SAAE's capability of optimizing the corresponding encoding distributions as expected. Besides, the discriminator can precisely predict the pure information components, which proves that SAAE is capable of learning to exclude the disparity components in latent codes through the adversarial training. The spectrum loss $L_{S}$ drops rapidly at first and then smoothly decreases until convergence over iterations.



\begin{figure}[htb]
    \centering 
    \resizebox{\linewidth}{!}{
\begin{subfigure}{0.44\linewidth}
  \centering
  \includegraphics[width=\textwidth]{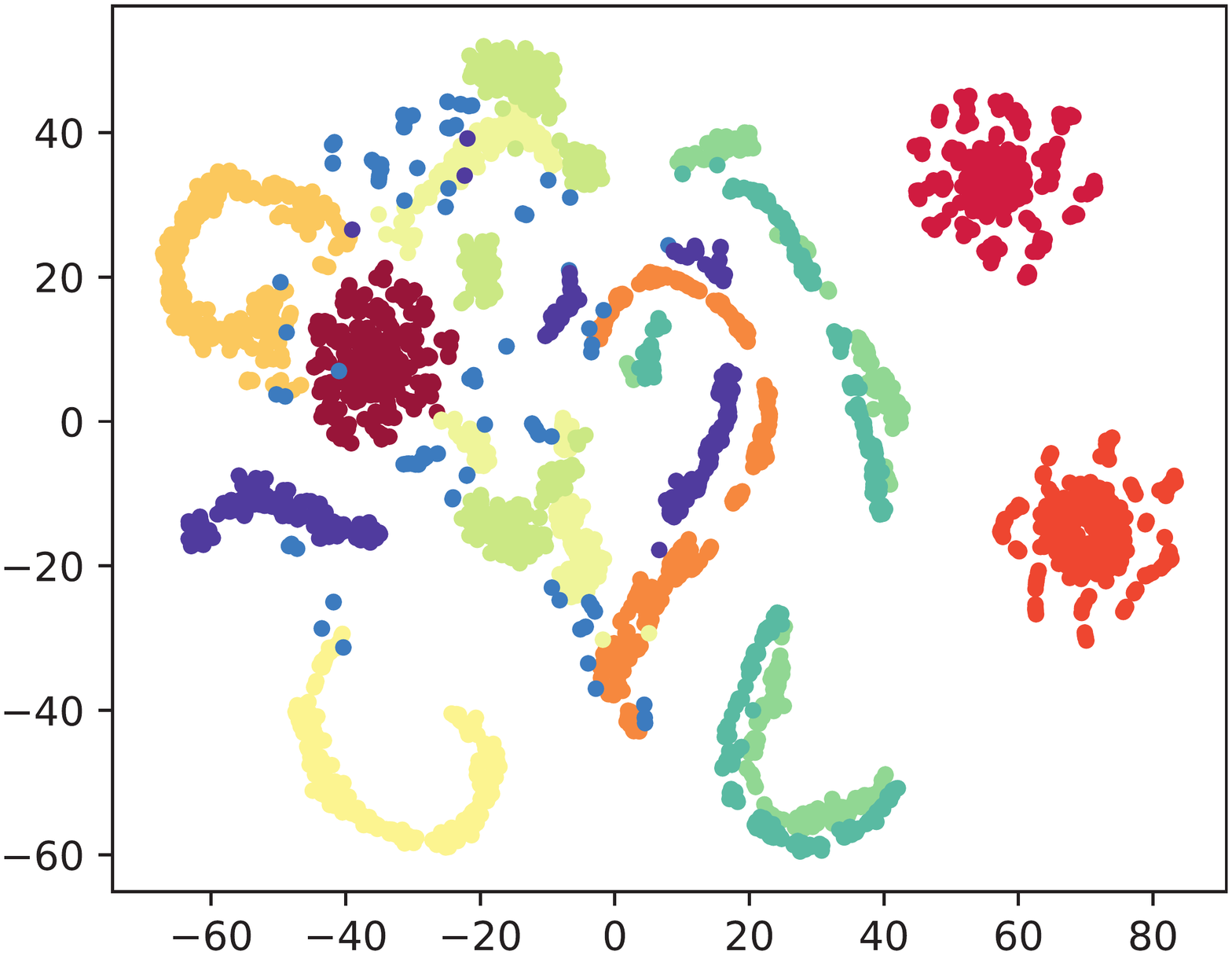}
  \caption{MHEALTH Raw}
\end{subfigure}\hfil 
\begin{subfigure}{0.44\linewidth}
  \includegraphics[width=\linewidth]{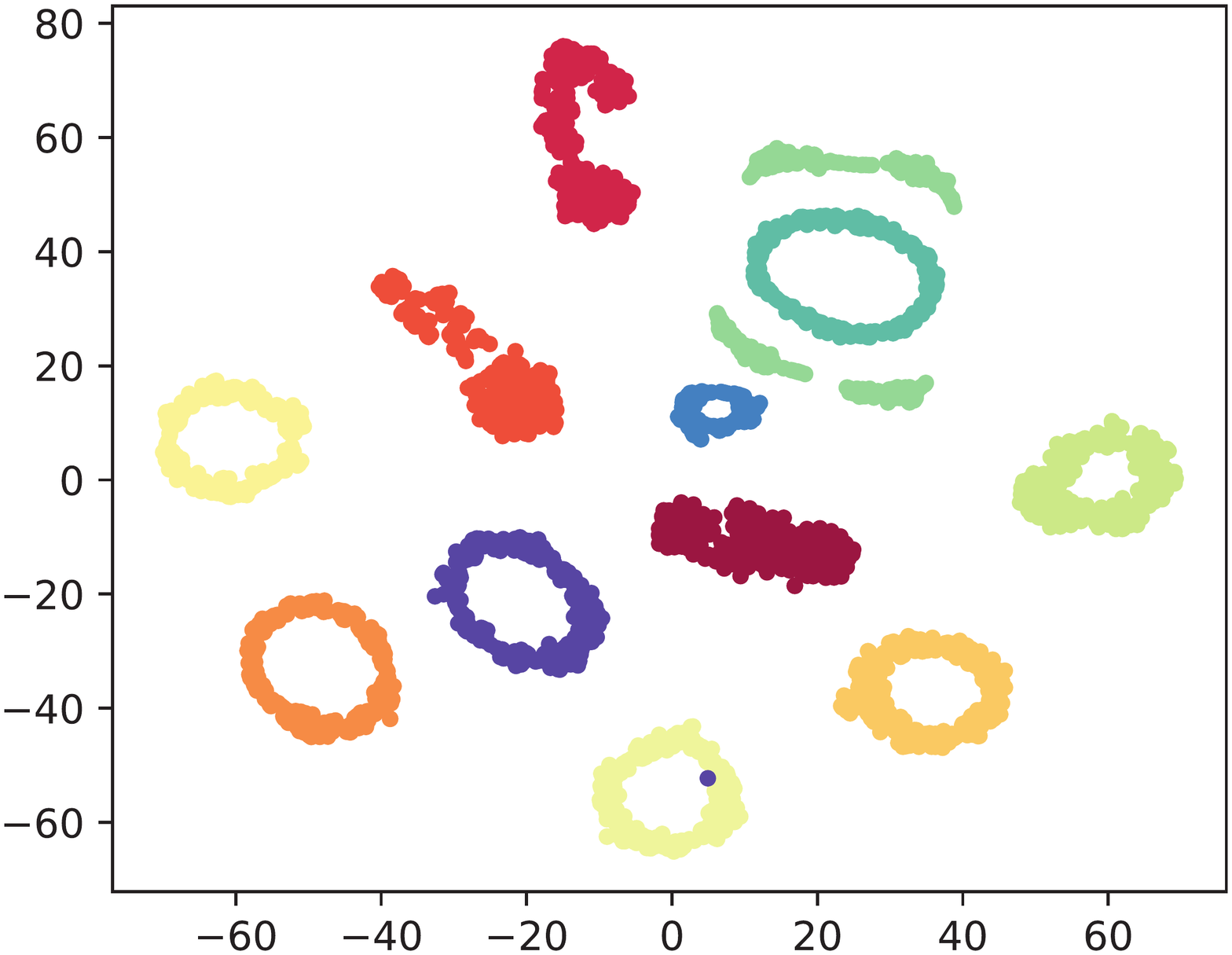}
    \centering
  \caption{MHEALTH SAAE}
\end{subfigure}\hfil 
}
\caption{Visualization of raw data and learned latent codes.}
\label{Embedding MHEALTH}
\end{figure}

\noindent\textbf{Visualization of Learned Representation.}
Fig. \ref{Embedding MHEALTH}  visualizes the raw data and the learned features. The original dimension is reduced by t-Distributed Stochastic Neighbor Embedding (t-SNE). The learned features are more similar within classes yet more dissimilar between classes over the sample space than the raw features, indicating the effectiveness of the network's feature learning.

\section{Related Work}
The previous research on intraclass disparity falls into two categories: subject-dependent and subject-independent models. Some of the subject-dependent algorithms attempt to discover patterns by case-specific analysis: Ren et al. \cite{activitysubjectvariance} captured the community in natural activities through hand trajectory; Tapia et al. \cite{activitysubjectvariance2} designed a more comprehensive system to record the invariant features of angle movement through wireless sensors. 

Another thread of subject-dependent research uses deep neural networks to learn representations for automated feature design and then activity recognition. Such methods focus on capturing the discriminative representations in the signal streams. Yang et al. \cite{mccnn} applied convolution and pooling filters along the temporal dimensions to catch the difference in the long-term time sequence signals; Fransisco et al. \cite{convlstm}, who further improved CNN with LSTM but still neglect the varying information amounts in signals. To validate the significance of signals, Vishvak et al. \cite{attention-convlstm} integrated a temporal attention module that aligns the output vector of the last time step with other steps' to learn a relative importance score to modify the learning process. However, subject-dependent methods fail to consider the unique patterns that may occur in new subjects, as the common patterns in existing subjects may not include all the potential conditions.

Subject-independent studies aim to enhance the generalization ability of models into precise recognition of new persons. Sani et al. \cite{sani2018matching} proposed to construct a support set and match the most similar instances to ease the unique patterns'. Yu et al. \cite{ensemblelstm} ensembled the models from different iterations and promise a more generalized model.
The above non-generative models still lack generalization due to the model limitation. Balabka et al. \cite{ae_HAR_2} applied AAE to approximate a generalized distribution to simulate human activity distributions. Thus, the encoder in AAE could be more generalized and robust when handling new subjects. Zhang et al.~\cite{ae_HAR} fused the advantages of variational autoencoders and generative adversarial networks and designed a regularized latent representations for generation. However, these works exclusively analyzed the time-domain information and all considered the intraclass disparity as meaningless noise and failed to extract the disparity distribution by adversarial training. Besides analyzing raw data patterns, some hand-engineered domain knowledge was introduced to further extract high-level discriminative information to assist predictions. The commonly used features include time-domain features (e.g., mean, variance, skewness) and frequency-domain features (e.g., power spectral density) \cite{janidarmian2017}. Some studies design new features containing temporal and structural information. For example, Hammerla et al. \cite{ecdf} proposed the Empirical Cumulative Density Function (ECDF) to extract the spatial information of the signal frames. Such methods are generally heuristic and lack generalization in different scenarios.


Our work differs from the studies above on two aspects: utilization of spectrum analysis and specific intraclass disparity distribution learning. We implement and embrace spectrum analysis as a tool in AAE to leverage the optimization based on signal information amount. Further, we specify and precisely portray the class-conditioned intraclass disparity in a learnable competitive encoding distribution, which enables AAE effectively to extract and to denoise such disparity.


\section{Conclusion}
We propose novel spectrum-guided disparity learning, or SAAE, to address intraclass variability. We design two competitive encoding distributions under a unified adversarial training framework, rather than a fixed prior distribution, to learn robust embeddings that can be generalized to new subjects.
We further incorporate the domain-specific knowledge in an unsupervised manner.
We experimentally validate our model on four representative benchmark dataset with state-of-the-art methods. The results demonstrate the superior performance and robustness of the proposed model in predictions on unknown subjects. Given SAAE's promising performance in handling intraclass disparity, we will extend it to handle more complex scenarios in the future.

\section{Acknowledgement}
This research was partially supported by grant ONRG NICOP N62909-19-1-2009.

\bibliographystyle{ACM-Reference-Format} 
\normalem
\bibliography{bio}


\begin{thebibliography}{30}


\ifx \showCODEN    \undefined \def \showCODEN     #1{\unskip}     \fi
\ifx \showDOI      \undefined \def \showDOI       #1{#1}\fi
\ifx \showISBNx    \undefined \def \showISBNx     #1{\unskip}     \fi
\ifx \showISBNxiii \undefined \def \showISBNxiii  #1{\unskip}     \fi
\ifx \showISSN     \undefined \def \showISSN      #1{\unskip}     \fi
\ifx \showLCCN     \undefined \def \showLCCN      #1{\unskip}     \fi
\ifx \shownote     \undefined \def \shownote      #1{#1}          \fi
\ifx \showarticletitle \undefined \def \showarticletitle #1{#1}   \fi
\ifx \showURL      \undefined \def \showURL       {\relax}        \fi
\providecommand\bibfield[2]{#2}
\providecommand\bibinfo[2]{#2}
\providecommand\natexlab[1]{#1}
\providecommand\showeprint[2][]{arXiv:#2}

\bibitem[\protect\citeauthoryear{Balabka}{Balabka}{2019}]%
        {ae_HAR_2}
\bibfield{author}{\bibinfo{person}{Dmitrijs Balabka}.}
  \bibinfo{year}{2019}\natexlab{}.
\newblock \showarticletitle{Semi-supervised learning for human activity
  recognition using adversarial autoencoders}. In
  \bibinfo{booktitle}{\emph{Proceedings of the 2019 ACM International Joint
  Conference on Pervasive and Ubiquitous Computing}}. ACM,
  \bibinfo{pages}{685--688}.
\newblock


\bibitem[\protect\citeauthoryear{Banos, Garcia, Holgado-Terriza, Damas,
  Pomares, Rojas, Saez, and Villalonga}{Banos et~al\mbox{.}}{2014}]%
        {MHEALTH}
\bibfield{author}{\bibinfo{person}{Oresti Banos}, \bibinfo{person}{Rafael
  Garcia}, \bibinfo{person}{Juan~A Holgado-Terriza}, \bibinfo{person}{Miguel
  Damas}, \bibinfo{person}{Hector Pomares}, \bibinfo{person}{Ignacio Rojas},
  \bibinfo{person}{Alejandro Saez}, {and} \bibinfo{person}{Claudia
  Villalonga}.} \bibinfo{year}{2014}\natexlab{}.
\newblock \showarticletitle{mHealthDroid: a novel framework for agile
  development of mobile health applications}. In
  \bibinfo{booktitle}{\emph{International workshop on ambient assisted
  living}}. Springer, \bibinfo{pages}{91--98}.
\newblock


\bibitem[\protect\citeauthoryear{Barshan and Y{\"u}ksek}{Barshan and
  Y{\"u}ksek}{2014}]%
        {ucidsads}
\bibfield{author}{\bibinfo{person}{Billur Barshan} {and}
  \bibinfo{person}{Murat~Cihan Y{\"u}ksek}.} \bibinfo{year}{2014}\natexlab{}.
\newblock \showarticletitle{Recognizing daily and sports activities in two open
  source machine learning environments using body-worn sensor units}.
\newblock \bibinfo{journal}{\emph{Comput. J.}} \bibinfo{volume}{57},
  \bibinfo{number}{11} (\bibinfo{year}{2014}), \bibinfo{pages}{1649--1667}.
\newblock


\bibitem[\protect\citeauthoryear{Cai, Chen, and Liang}{Cai
  et~al\mbox{.}}{2015}]%
        {intra_in_face_recoginition}
\bibfield{author}{\bibinfo{person}{Jun Cai}, \bibinfo{person}{Jing Chen}, {and}
  \bibinfo{person}{Xing Liang}.} \bibinfo{year}{2015}\natexlab{}.
\newblock \showarticletitle{Single-sample face recognition based on intra-class
  differences in a variation model}.
\newblock \bibinfo{journal}{\emph{Sensors}} \bibinfo{volume}{15},
  \bibinfo{number}{1} (\bibinfo{year}{2015}), \bibinfo{pages}{1071--1087}.
\newblock


\bibitem[\protect\citeauthoryear{Chen, Zhang, Yao, Guo, Yu, and Liu}{Chen
  et~al\mbox{.}}{2020}]%
        {chen2020deep}
\bibfield{author}{\bibinfo{person}{Kaixuan Chen}, \bibinfo{person}{Dalin
  Zhang}, \bibinfo{person}{Lina Yao}, \bibinfo{person}{Bin Guo},
  \bibinfo{person}{Zhiwen Yu}, {and} \bibinfo{person}{Yunhao Liu}.}
  \bibinfo{year}{2020}\natexlab{}.
\newblock \showarticletitle{Deep learning for sensor-based human activity
  recognition: overview, challenges and opportunities}.
\newblock \bibinfo{journal}{\emph{arXiv preprint arXiv:2001.07416}}
  (\bibinfo{year}{2020}).
\newblock


\bibitem[\protect\citeauthoryear{Creswell and Bharath}{Creswell and
  Bharath}{2018}]%
        {creswell2018denoising}
\bibfield{author}{\bibinfo{person}{Antonia Creswell} {and}
  \bibinfo{person}{Anil~Anthony Bharath}.} \bibinfo{year}{2018}\natexlab{}.
\newblock \showarticletitle{Denoising adversarial autoencoders}.
\newblock \bibinfo{journal}{\emph{IEEE transactions on neural networks and
  learning systems}} \bibinfo{volume}{30}, \bibinfo{number}{4}
  (\bibinfo{year}{2018}), \bibinfo{pages}{968--984}.
\newblock


\bibitem[\protect\citeauthoryear{Dong-DongChen and WeiGao}{Dong-DongChen and
  WeiGao}{2018}]%
        {dong2018tri}
\bibfield{author}{\bibinfo{person}{WeiWang Dong-DongChen} {and}
  \bibinfo{person}{Zhi-HuaZhou WeiGao}.} \bibinfo{year}{2018}\natexlab{}.
\newblock \showarticletitle{Tri-net for semi-supervised deep learning}. In
  \bibinfo{booktitle}{\emph{Proceedings of Twenty-Seventh International Joint
  Conference on Artificial Intelligence}}. \bibinfo{pages}{2014--2020}.
\newblock


\bibitem[\protect\citeauthoryear{Goodfellow, Pouget-Abadie, Mirza, Xu,
  Warde-Farley, Ozair, Courville, and Bengio}{Goodfellow et~al\mbox{.}}{2014}]%
        {GAN}
\bibfield{author}{\bibinfo{person}{Ian Goodfellow}, \bibinfo{person}{Jean
  Pouget-Abadie}, \bibinfo{person}{Mehdi Mirza}, \bibinfo{person}{Bing Xu},
  \bibinfo{person}{David Warde-Farley}, \bibinfo{person}{Sherjil Ozair},
  \bibinfo{person}{Aaron Courville}, {and} \bibinfo{person}{Yoshua Bengio}.}
  \bibinfo{year}{2014}\natexlab{}.
\newblock \showarticletitle{Generative adversarial nets}. In
  \bibinfo{booktitle}{\emph{Advances in neural information processing
  systems}}. \bibinfo{pages}{2672--2680}.
\newblock


\bibitem[\protect\citeauthoryear{Guan and Pl{\"o}tz}{Guan and
  Pl{\"o}tz}{2017}]%
        {ensemblelstm}
\bibfield{author}{\bibinfo{person}{Yu Guan} {and} \bibinfo{person}{Thomas
  Pl{\"o}tz}.} \bibinfo{year}{2017}\natexlab{}.
\newblock \showarticletitle{Ensembles of deep lstm learners for activity
  recognition using wearables}.
\newblock \bibinfo{journal}{\emph{Proceedings of the ACM on Interactive,
  Mobile, Wearable and Ubiquitous Technologies}} \bibinfo{volume}{1},
  \bibinfo{number}{2} (\bibinfo{year}{2017}), \bibinfo{pages}{11}.
\newblock


\bibitem[\protect\citeauthoryear{Hammerla, Halloran, and Pl{\"o}tz}{Hammerla
  et~al\mbox{.}}{2016}]%
        {2016ijcai}
\bibfield{author}{\bibinfo{person}{Nils~Y Hammerla}, \bibinfo{person}{Shane
  Halloran}, {and} \bibinfo{person}{Thomas Pl{\"o}tz}.}
  \bibinfo{year}{2016}\natexlab{}.
\newblock \showarticletitle{Deep, convolutional, and recurrent models for human
  activity recognition using wearables}. In
  \bibinfo{booktitle}{\emph{Proceedings of the Twenty-Fifth International Joint
  Conference on Artificial Intelligence}}. AAAI Press,
  \bibinfo{pages}{1533--1540}.
\newblock


\bibitem[\protect\citeauthoryear{Hammerla, Kirkham, Andras, and
  Ploetz}{Hammerla et~al\mbox{.}}{2013}]%
        {ecdf}
\bibfield{author}{\bibinfo{person}{Nils~Y Hammerla}, \bibinfo{person}{Reuben
  Kirkham}, \bibinfo{person}{Peter Andras}, {and} \bibinfo{person}{Thomas
  Ploetz}.} \bibinfo{year}{2013}\natexlab{}.
\newblock \showarticletitle{On preserving statistical characteristics of
  accelerometry data using their empirical cumulative distribution}. In
  \bibinfo{booktitle}{\emph{Proceedings of the 2013 International Symposium on
  Wearable Computers}}. ACM, \bibinfo{pages}{65--68}.
\newblock


\bibitem[\protect\citeauthoryear{Im, Ahn, Memisevic, and Bengio}{Im
  et~al\mbox{.}}{2017}]%
        {im2017denoising}
\bibfield{author}{\bibinfo{person}{Daniel Im~Jiwoong Im},
  \bibinfo{person}{Sungjin Ahn}, \bibinfo{person}{Roland Memisevic}, {and}
  \bibinfo{person}{Yoshua Bengio}.} \bibinfo{year}{2017}\natexlab{}.
\newblock \showarticletitle{Denoising criterion for variational auto-encoding
  framework}. In \bibinfo{booktitle}{\emph{Thirty-First AAAI Conference on
  Artificial Intelligence}}.
\newblock


\bibitem[\protect\citeauthoryear{Janidarmian, Roshan~Fekr, Radecka, and
  Zilic}{Janidarmian et~al\mbox{.}}{2017}]%
        {janidarmian2017}
\bibfield{author}{\bibinfo{person}{Majid Janidarmian}, \bibinfo{person}{Atena
  Roshan~Fekr}, \bibinfo{person}{Katarzyna Radecka}, {and}
  \bibinfo{person}{Zeljko Zilic}.} \bibinfo{year}{2017}\natexlab{}.
\newblock \showarticletitle{A comprehensive analysis on wearable acceleration
  sensors in human activity recognition}.
\newblock \bibinfo{journal}{\emph{Sensors}} \bibinfo{volume}{17},
  \bibinfo{number}{3} (\bibinfo{year}{2017}), \bibinfo{pages}{529}.
\newblock


\bibitem[\protect\citeauthoryear{Kingma and Ba}{Kingma and Ba}{2014}]%
        {kingma2014adam}
\bibfield{author}{\bibinfo{person}{Diederik~P Kingma} {and}
  \bibinfo{person}{Jimmy Ba}.} \bibinfo{year}{2014}\natexlab{}.
\newblock \showarticletitle{Adam: A method for stochastic optimization}.
\newblock \bibinfo{journal}{\emph{arXiv preprint arXiv:1412.6980}}
  (\bibinfo{year}{2014}).
\newblock


\bibitem[\protect\citeauthoryear{Kingma and Welling}{Kingma and
  Welling}{2013}]%
        {VAE}
\bibfield{author}{\bibinfo{person}{Diederik~P Kingma} {and}
  \bibinfo{person}{Max Welling}.} \bibinfo{year}{2013}\natexlab{}.
\newblock \showarticletitle{Auto-encoding variational bayes}.
\newblock \bibinfo{journal}{\emph{arXiv preprint arXiv:1312.6114}}
  (\bibinfo{year}{2013}).
\newblock


\bibitem[\protect\citeauthoryear{Makhzani, Shlens, Jaitly, Goodfellow, and
  Frey}{Makhzani et~al\mbox{.}}{2015}]%
        {adversarialautoencoder}
\bibfield{author}{\bibinfo{person}{Alireza Makhzani}, \bibinfo{person}{Jonathon
  Shlens}, \bibinfo{person}{Navdeep Jaitly}, \bibinfo{person}{Ian Goodfellow},
  {and} \bibinfo{person}{Brendan Frey}.} \bibinfo{year}{2015}\natexlab{}.
\newblock \showarticletitle{Adversarial autoencoders}.
\newblock \bibinfo{journal}{\emph{arXiv preprint arXiv:1511.05644}}
  (\bibinfo{year}{2015}).
\newblock


\bibitem[\protect\citeauthoryear{Murahari and Pl{\"o}tz}{Murahari and
  Pl{\"o}tz}{2018}]%
        {attention-convlstm}
\bibfield{author}{\bibinfo{person}{Vishvak~S Murahari} {and}
  \bibinfo{person}{Thomas Pl{\"o}tz}.} \bibinfo{year}{2018}\natexlab{}.
\newblock \showarticletitle{On attention models for human activity
  recognition}. In \bibinfo{booktitle}{\emph{Proceedings of the 2018 ACM
  International Symposium on Wearable Computers}}. ACM,
  \bibinfo{pages}{100--103}.
\newblock


\bibitem[\protect\citeauthoryear{Ord{\'o}{\~n}ez and Roggen}{Ord{\'o}{\~n}ez
  and Roggen}{2016}]%
        {convlstm}
\bibfield{author}{\bibinfo{person}{Francisco Ord{\'o}{\~n}ez} {and}
  \bibinfo{person}{Daniel Roggen}.} \bibinfo{year}{2016}\natexlab{}.
\newblock \showarticletitle{Deep convolutional and lstm recurrent neural
  networks for multimodal wearable activity recognition}.
\newblock \bibinfo{journal}{\emph{Sensors}} \bibinfo{volume}{16},
  \bibinfo{number}{1} (\bibinfo{year}{2016}), \bibinfo{pages}{115}.
\newblock


\bibitem[\protect\citeauthoryear{Reiss and Stricker}{Reiss and
  Stricker}{2012}]%
        {pamap2}
\bibfield{author}{\bibinfo{person}{Attila Reiss} {and} \bibinfo{person}{Didier
  Stricker}.} \bibinfo{year}{2012}\natexlab{}.
\newblock \showarticletitle{Introducing a new benchmarked dataset for activity
  monitoring}. In \bibinfo{booktitle}{\emph{2012 16th International Symposium
  on Wearable Computers}}. IEEE, \bibinfo{pages}{108--109}.
\newblock


\bibitem[\protect\citeauthoryear{Ren, Xu, and Kee}{Ren et~al\mbox{.}}{2004}]%
        {activitysubjectvariance2}
\bibfield{author}{\bibinfo{person}{Haibing Ren}, \bibinfo{person}{Guangyou Xu},
  {and} \bibinfo{person}{SeokCheol Kee}.} \bibinfo{year}{2004}\natexlab{}.
\newblock \showarticletitle{Subject-independent natural action recognition}. In
  \bibinfo{booktitle}{\emph{Sixth IEEE International Conference on Automatic
  Face and Gesture Recognition, 2004. Proceedings.}} IEEE,
  \bibinfo{pages}{523--528}.
\newblock


\bibitem[\protect\citeauthoryear{Roggen, Calatroni, Rossi, Holleczek,
  et~al\mbox{.}}{Roggen et~al\mbox{.}}{2010}]%
        {oppo}
\bibfield{author}{\bibinfo{person}{Daniel Roggen}, \bibinfo{person}{Alberto
  Calatroni}, \bibinfo{person}{Mirco Rossi}, \bibinfo{person}{Holleczek},
  {et~al\mbox{.}}} \bibinfo{year}{2010}\natexlab{}.
\newblock \showarticletitle{Collecting complex activity datasets in highly rich
  networked sensor environments}. In \bibinfo{booktitle}{\emph{The 7th
  International conference on networked sensing systems (INSS)}}. IEEE,
  \bibinfo{pages}{233--240}.
\newblock


\bibitem[\protect\citeauthoryear{Sani, Wiratunga, Massie, and Cooper}{Sani
  et~al\mbox{.}}{2018}]%
        {sani2018matching}
\bibfield{author}{\bibinfo{person}{Sadiq Sani}, \bibinfo{person}{Nirmalie
  Wiratunga}, \bibinfo{person}{Stewart Massie}, {and} \bibinfo{person}{Kay
  Cooper}.} \bibinfo{year}{2018}\natexlab{}.
\newblock \showarticletitle{Matching networks for personalised human activity
  recognition.} CEUR Workshop Proceedings.
\newblock


\bibitem[\protect\citeauthoryear{Smith et~al\mbox{.}}{Smith
  et~al\mbox{.}}{1997}]%
        {DFT}
\bibfield{author}{\bibinfo{person}{Steven~W Smith} {et~al\mbox{.}}}
  \bibinfo{year}{1997}\natexlab{}.
\newblock \showarticletitle{The scientist and engineer's guide to digital
  signal processing}.
\newblock  (\bibinfo{year}{1997}).
\newblock


\bibitem[\protect\citeauthoryear{Suryadevara and Mukhopadhyay}{Suryadevara and
  Mukhopadhyay}{2014}]%
        {healthcare}
\bibfield{author}{\bibinfo{person}{Nagender~K Suryadevara} {and}
  \bibinfo{person}{Subhas~C Mukhopadhyay}.} \bibinfo{year}{2014}\natexlab{}.
\newblock \showarticletitle{Determining wellness through an ambient assisted
  living environment}.
\newblock \bibinfo{journal}{\emph{IEEE Intelligent Systems}}
  \bibinfo{volume}{29}, \bibinfo{number}{3} (\bibinfo{year}{2014}),
  \bibinfo{pages}{30--37}.
\newblock


\bibitem[\protect\citeauthoryear{Tapia, Intille, Haskell, and et~al}{Tapia
  et~al\mbox{.}}{2007}]%
        {activitysubjectvariance}
\bibfield{author}{\bibinfo{person}{Emmanuel~Munguia Tapia},
  \bibinfo{person}{Stephen~S Intille}, \bibinfo{person}{William Haskell}, {and}
  \bibinfo{person}{Larson et al}.} \bibinfo{year}{2007}\natexlab{}.
\newblock \showarticletitle{Real-time recognition of physical activities and
  their intensities using wireless accelerometers and a heart rate monitor}. In
  \bibinfo{booktitle}{\emph{The 11th IEEE international symposium on wearable
  computers}}. IEEE, \bibinfo{pages}{37--40}.
\newblock


\bibitem[\protect\citeauthoryear{Vincent, Larochelle, Bengio, and
  Manzagol}{Vincent et~al\mbox{.}}{2008}]%
        {vincent2008extracting}
\bibfield{author}{\bibinfo{person}{Pascal Vincent}, \bibinfo{person}{Hugo
  Larochelle}, \bibinfo{person}{Yoshua Bengio}, {and}
  \bibinfo{person}{Pierre-Antoine Manzagol}.} \bibinfo{year}{2008}\natexlab{}.
\newblock \showarticletitle{Extracting and composing robust features with
  denoising autoencoders}. In \bibinfo{booktitle}{\emph{Proceedings of the 25th
  international conference on Machine learning}}. ACM,
  \bibinfo{pages}{1096--1103}.
\newblock


\bibitem[\protect\citeauthoryear{Viterbi and Omura}{Viterbi and Omura}{2013}]%
        {signal_Principle}
\bibfield{author}{\bibinfo{person}{Andrew~J Viterbi} {and}
  \bibinfo{person}{Jim~K Omura}.} \bibinfo{year}{2013}\natexlab{}.
\newblock \bibinfo{booktitle}{\emph{Principles of digital communication and
  coding}}.
\newblock \bibinfo{publisher}{Courier Corporation}.
\newblock


\bibitem[\protect\citeauthoryear{Yang, Nguyen, San, Li, and Krishnaswamy}{Yang
  et~al\mbox{.}}{2015}]%
        {mccnn}
\bibfield{author}{\bibinfo{person}{Jianbo Yang}, \bibinfo{person}{Minh~Nhut
  Nguyen}, \bibinfo{person}{Phyo~Phyo San}, \bibinfo{person}{Xiao~Li Li}, {and}
  \bibinfo{person}{Shonali Krishnaswamy}.} \bibinfo{year}{2015}\natexlab{}.
\newblock \showarticletitle{Deep convolutional neural networks on multichannel
  time series for human activity recognition}. In
  \bibinfo{booktitle}{\emph{Twenty-Fourth International Joint Conference on
  Artificial Intelligence}}.
\newblock


\bibitem[\protect\citeauthoryear{Yao, Sheng, Li, and et~all}{Yao
  et~al\mbox{.}}{2017}]%
        {yao2017compressive}
\bibfield{author}{\bibinfo{person}{Lina Yao}, \bibinfo{person}{Quan~Z Sheng},
  \bibinfo{person}{Xue Li}, {and} \bibinfo{person}{Li et all}.}
  \bibinfo{year}{2017}\natexlab{}.
\newblock \showarticletitle{Compressive representation for device-free activity
  recognition with passive RFID signal strength}.
\newblock \bibinfo{journal}{\emph{IEEE Transactions on Mobile Computing (TMC)}}
  \bibinfo{volume}{17}, \bibinfo{number}{2} (\bibinfo{year}{2017}),
  \bibinfo{pages}{293--306}.
\newblock


\bibitem[\protect\citeauthoryear{Zhang, Yao, and Yuan}{Zhang
  et~al\mbox{.}}{2019}]%
        {ae_HAR}
\bibfield{author}{\bibinfo{person}{Xiang Zhang}, \bibinfo{person}{Lina Yao},
  {and} \bibinfo{person}{Feng Yuan}.} \bibinfo{year}{2019}\natexlab{}.
\newblock \showarticletitle{Adversarial Variational Embedding for Robust
  Semi-supervised Learning}. In \bibinfo{booktitle}{\emph{Proceedings of the
  25th ACM SIGKDD International Conference on Knowledge Discovery \& Data
  Mining}}.
\newblock


\end{thebibliography}
\newpage
\appendix
\section{Network Architecture}\label{Architecture}

\begin{figure}[htb]
    \centering
    \includegraphics[width=0.8\linewidth]{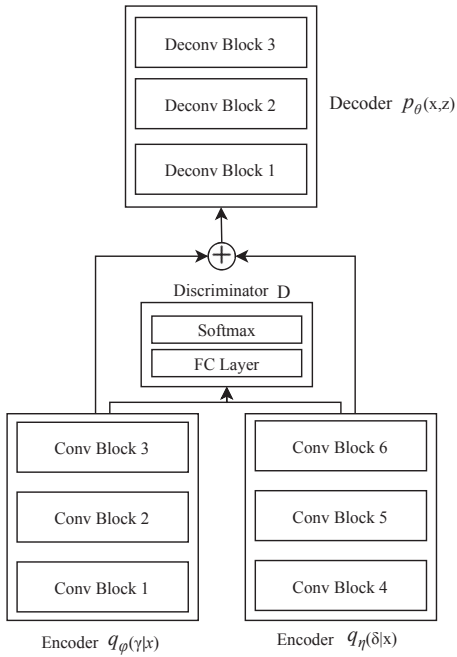}
    \caption{Network Detail}
    \label{Network Details}
\end{figure}

\begin{figure}[htb]
    \centering
    \includegraphics[width=\linewidth]{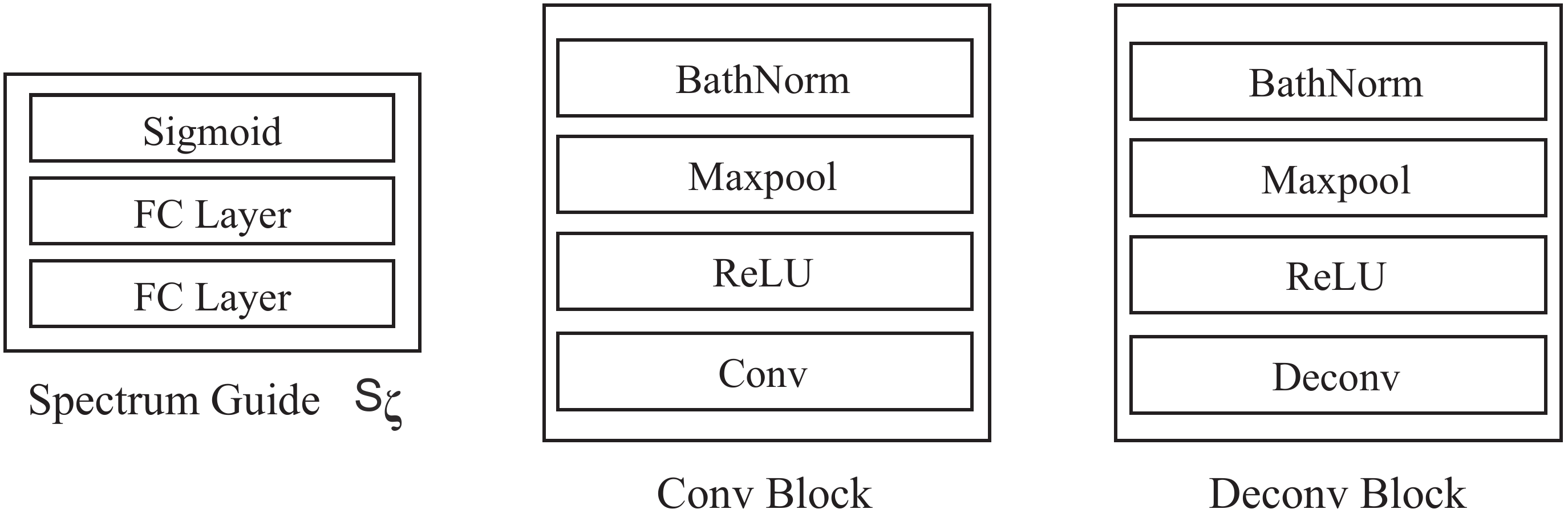}
    \caption{Module Detail}
    \label{Module Details}
\end{figure}

We plot the detailed network structure in Fig. \ref{Network Details}. The decoder and encoders consist of three independent blocks and two encoders share the same structure. The architectures of Block are shown in Fig. \ref{Module Details} and the parameters are shown in Table \ref{archi}. Specially, Conv Block 3 \& 6 and Deconv Block 3 do not have Maxpool layer. The stride of layers are 1 and padding way is 0 as default. Discriminator $D$ is composed of one FC layer, so the input dim equals the output element number of encoders and output dim equals the target class number. Given a amplitude spectrum $A\in \mathbb{R}^{1\times m}$, then the Spectrum Guide Function $S_{\zeta}$ consists of two FC layers: first layer takes $2\cdot m$ dimension input (i.e., $A^{N}$ and $A^{O}$ and keeps the same output dimension; second layer generates $m$-dimension outputs, which represents the corresponding weights for the frequencies in spectrum.

\begin{table}[t]
\centering
\caption{Block Parameters.}
\small
\smallskip
\begin{tabular}{c|cccc}
  Name & \#Kernel & Kernel Shape & Maxpool \\
  \hline
   Conv Block 1\&4  & 50 & (5,1) & (2,1) \\
   Conv Block 2\&5  & 40 & (5,1) & (2,1) \\
   Conv Block 3\&6  & 20 & (2,1) & none \\
   \hline
   Deconv Block 1  & 40 & (5,1) & (2,1) \\
   Deconv Block 2  & 50 & (5,1) & (2,1) \\
   Deconv Block 3  & 2 & (2,1) & none \\
\end{tabular}
\label{archi}
\end{table}
\section{Supplementary Experiment}
\begin{figure*}[htb]
    \centering 
\begin{subfigure}{0.24\textwidth}
  \includegraphics[width=\linewidth]{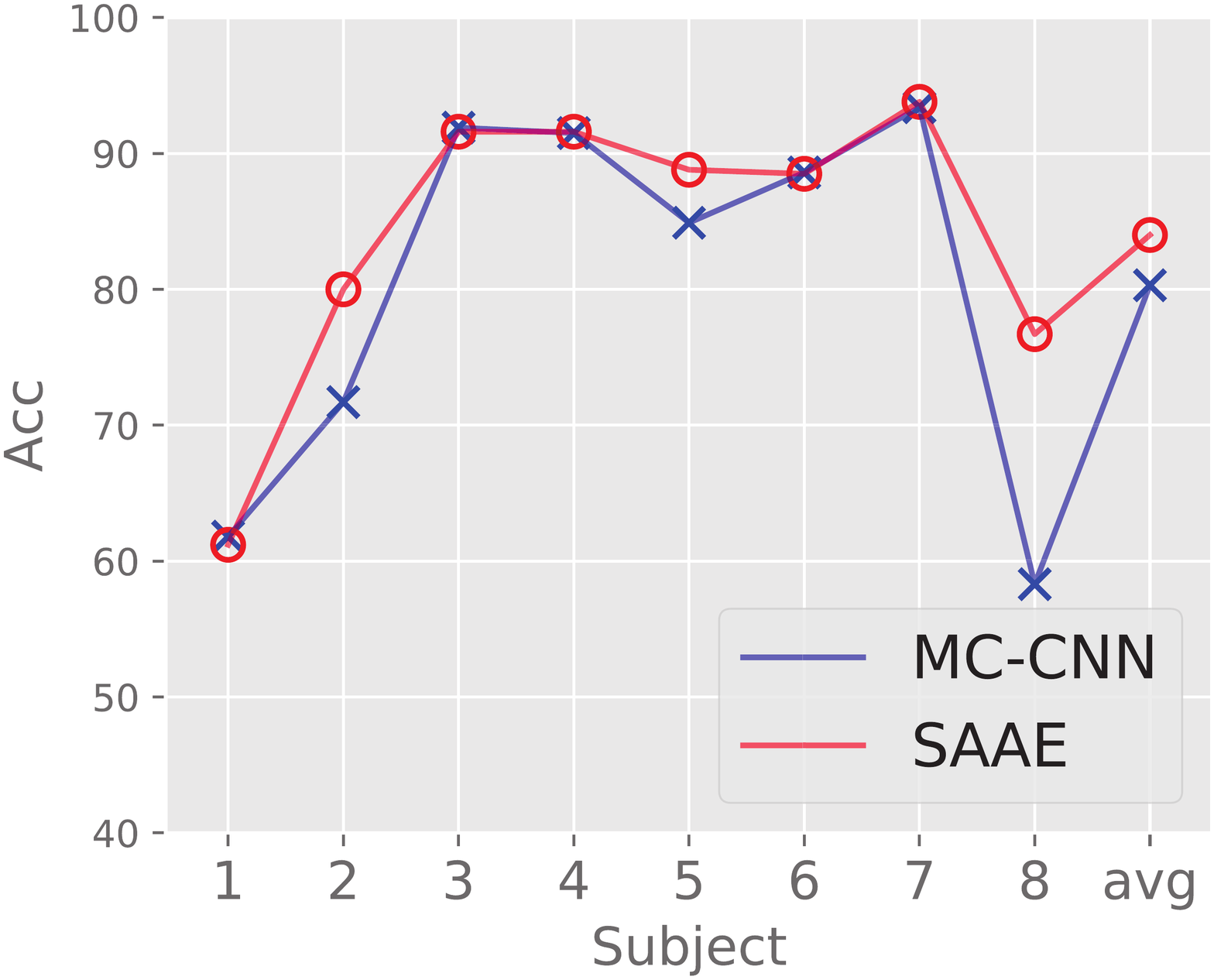}
    \centering
  \caption{PAMAP2 Acc Report}
\end{subfigure}\hfil 
\hspace{-1mm}
\begin{subfigure}{0.24\textwidth}
  \includegraphics[width=\linewidth]{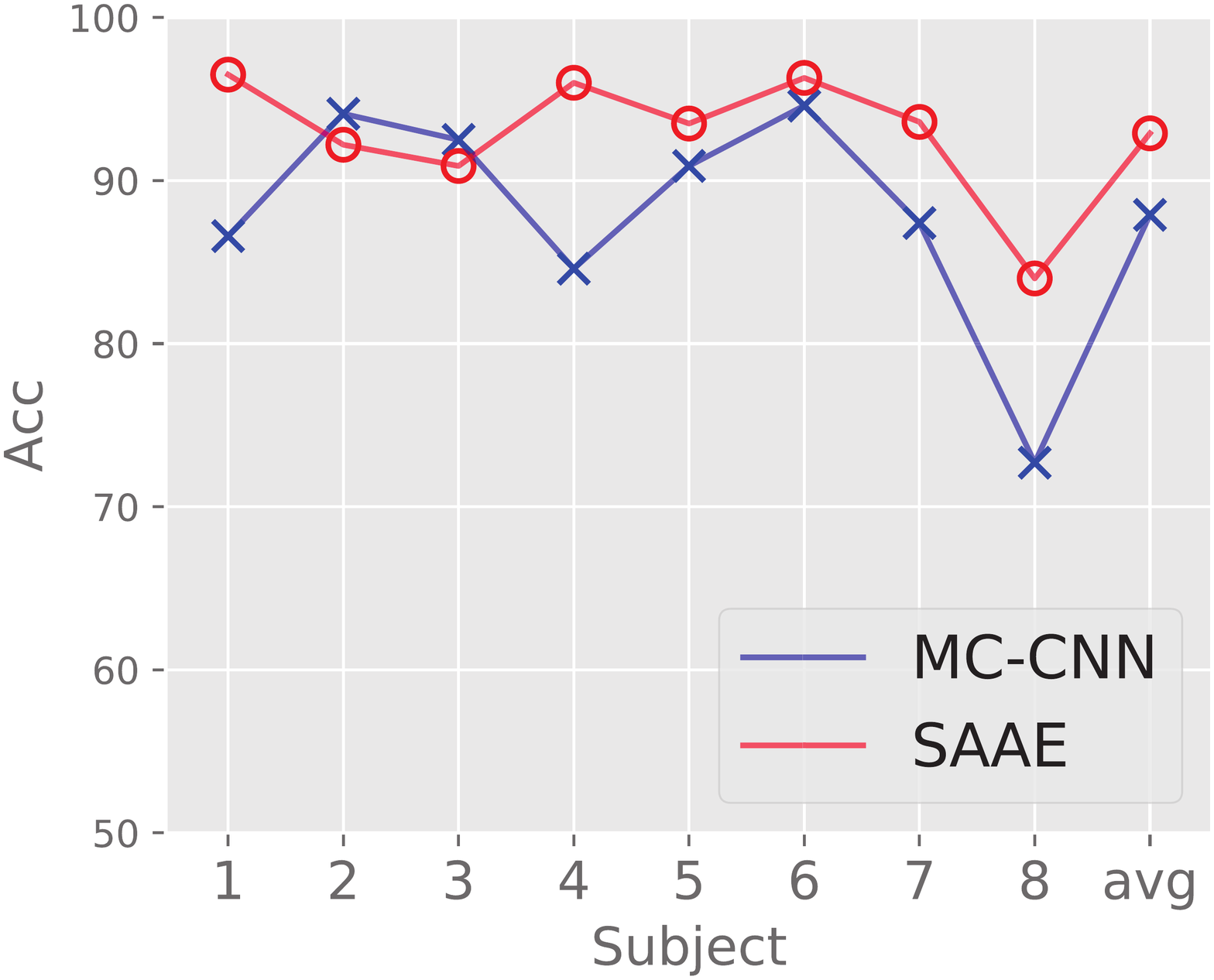}
    \centering
  \caption{UCIDSADS Acc Report}
\end{subfigure}
\hspace{-1mm}
\begin{subfigure}{0.24\textwidth}
  \includegraphics[width=\linewidth]{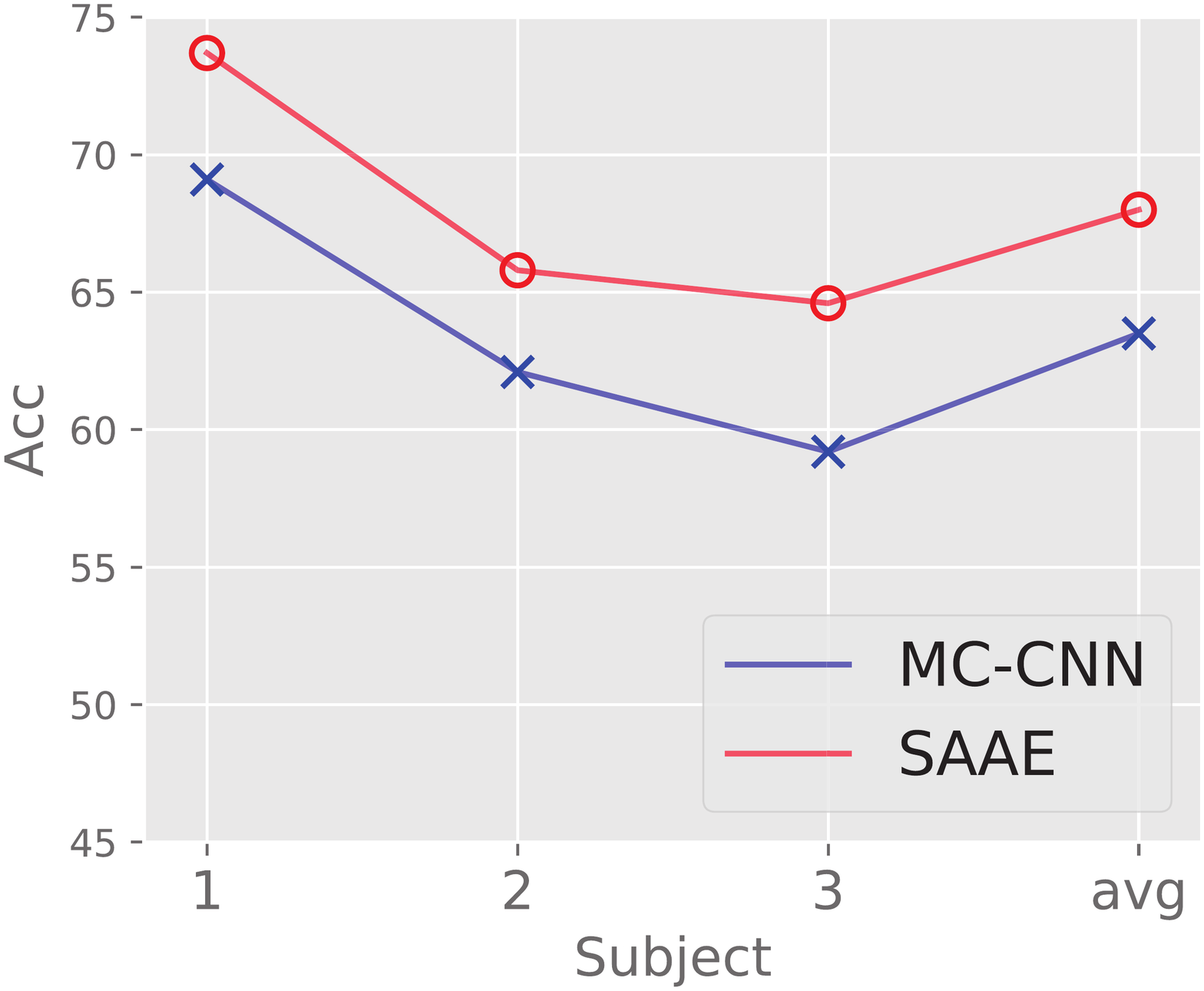}
    \centering
  \caption{OPPORTUNITY Acc Report}
\end{subfigure}
\hspace{-1mm}
\begin{subfigure}{0.24\textwidth}
  \includegraphics[width=\linewidth]{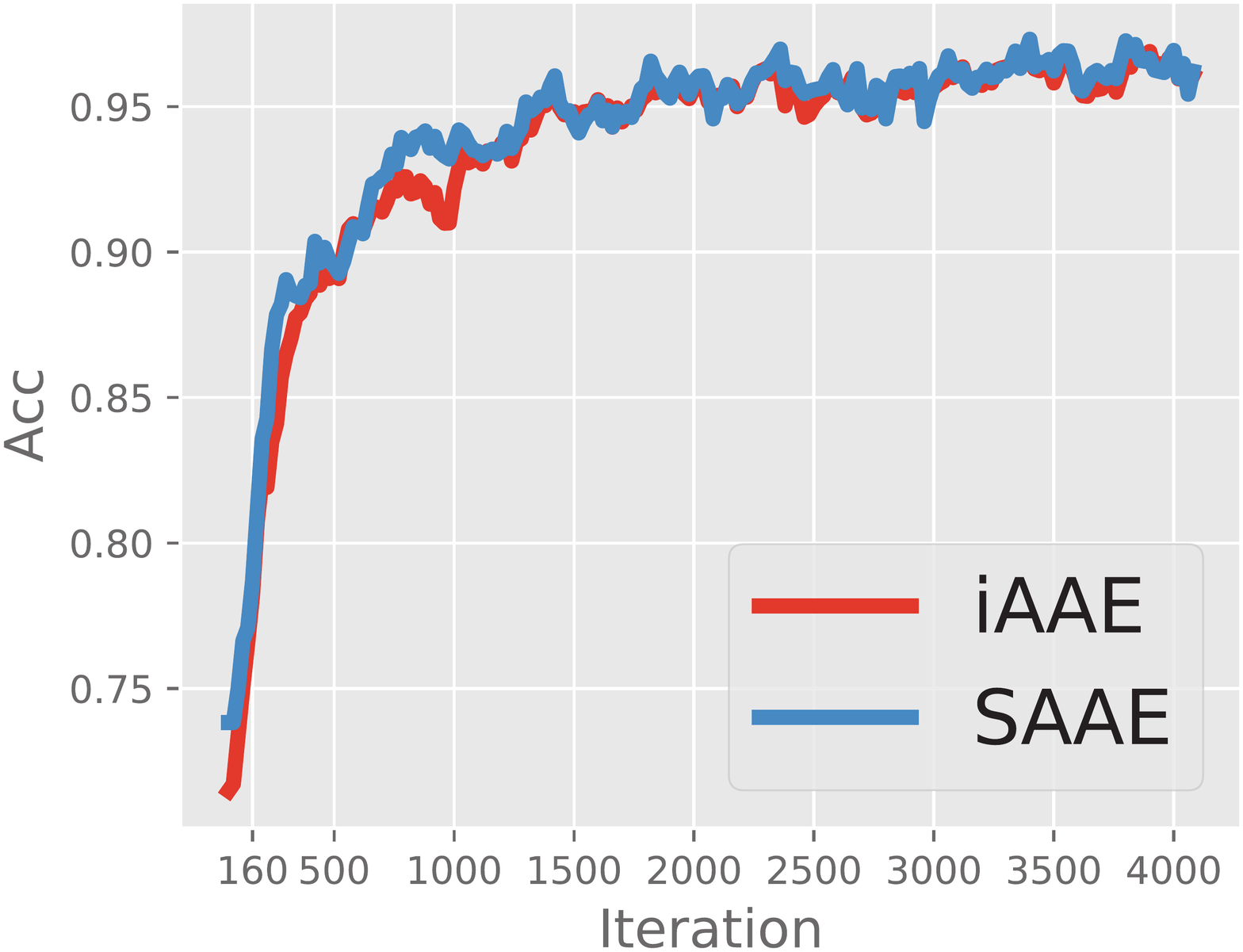}
    \centering
  \caption{\quad PAMAP2 Dis Acc}
\end{subfigure}\hfil 
\hspace{-1mm}
\begin{subfigure}{0.24\textwidth}
  \includegraphics[width=\linewidth]{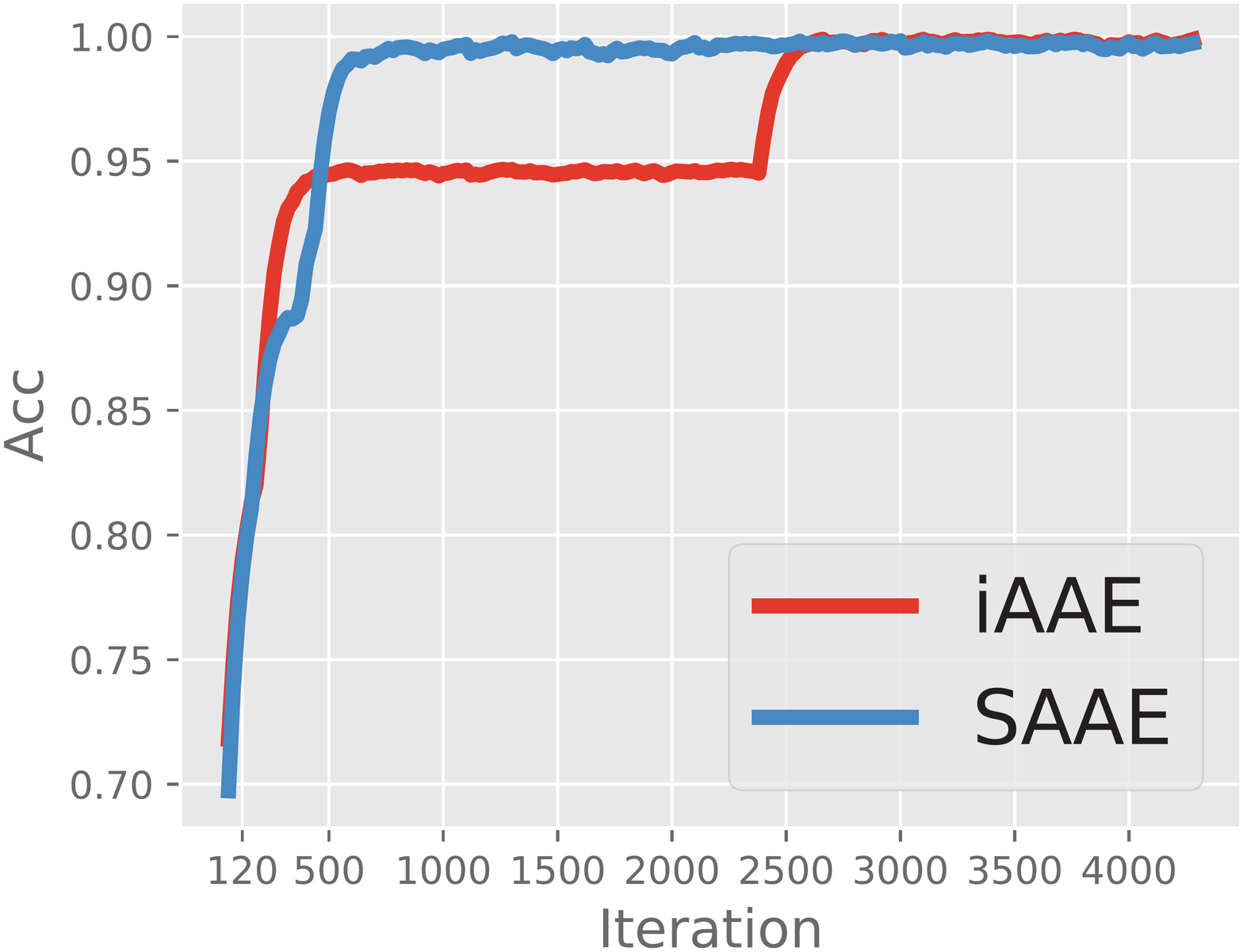}
    \centering
  \caption{\quad UCIDSADS Dis Acc}
\end{subfigure}
\hspace{-1mm}
\begin{subfigure}{0.24\textwidth}
  \includegraphics[width=\linewidth]{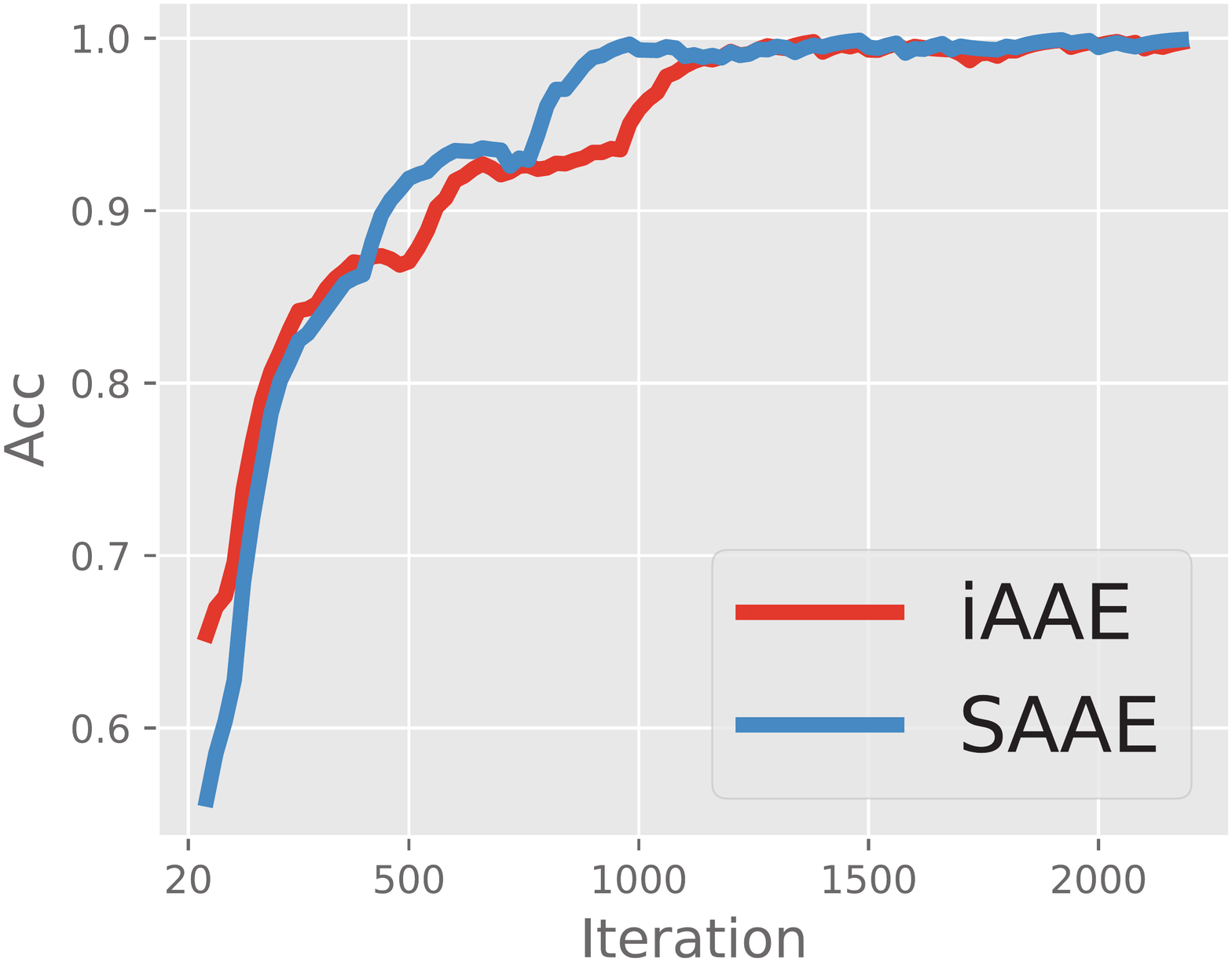}
    \centering
  \caption{\quad OPPORTUNITY Dis Acc}
\end{subfigure}
\begin{subfigure}{0.24\textwidth}
  \centering
  \includegraphics[width=\textwidth]{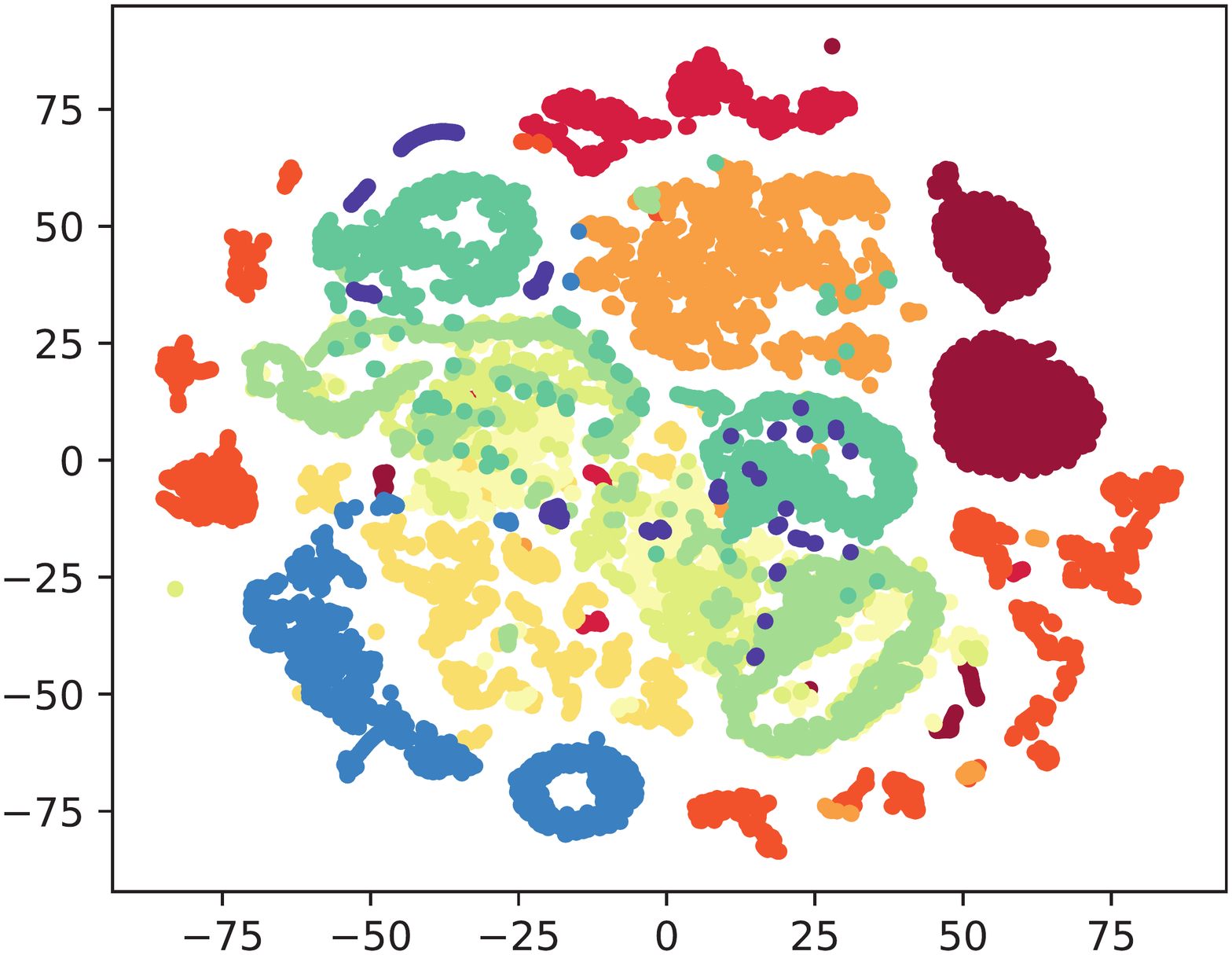}
  \caption{PAMAP2 Raw}
\end{subfigure}\hfil 
\hspace{-1mm}
\begin{subfigure}{0.24\textwidth}
  \includegraphics[width=\linewidth]{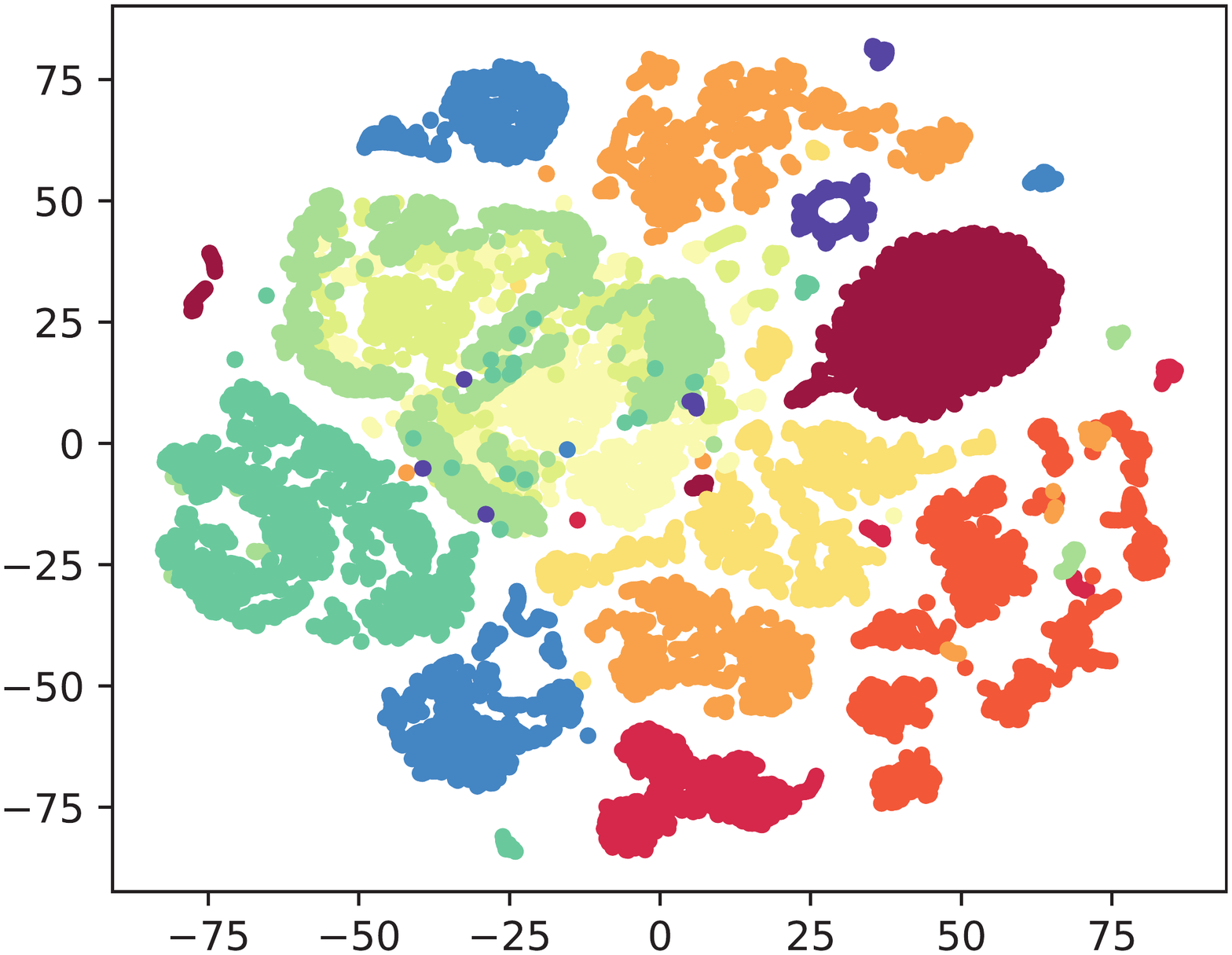}
    \centering
  \caption{PAMAP2 SAAE}
\end{subfigure}\hfil 
\hspace{-1mm}
\begin{subfigure}{0.24\textwidth}
  \centering
  \includegraphics[width=\textwidth]{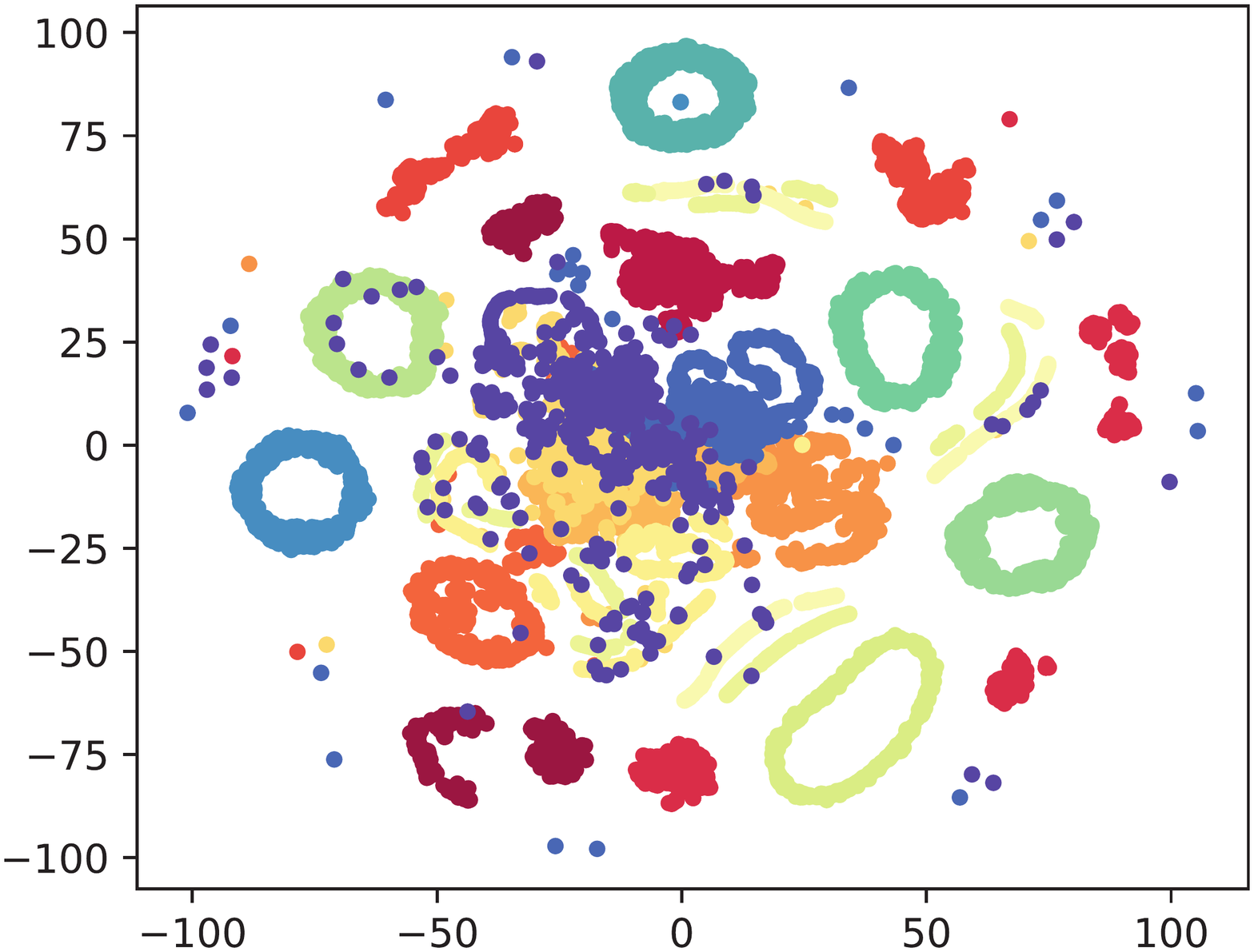}
  \caption{UCIDSADS Raw}
\end{subfigure}
\hspace{-1mm}
\begin{subfigure}{0.24\textwidth}
  \includegraphics[width=\linewidth]{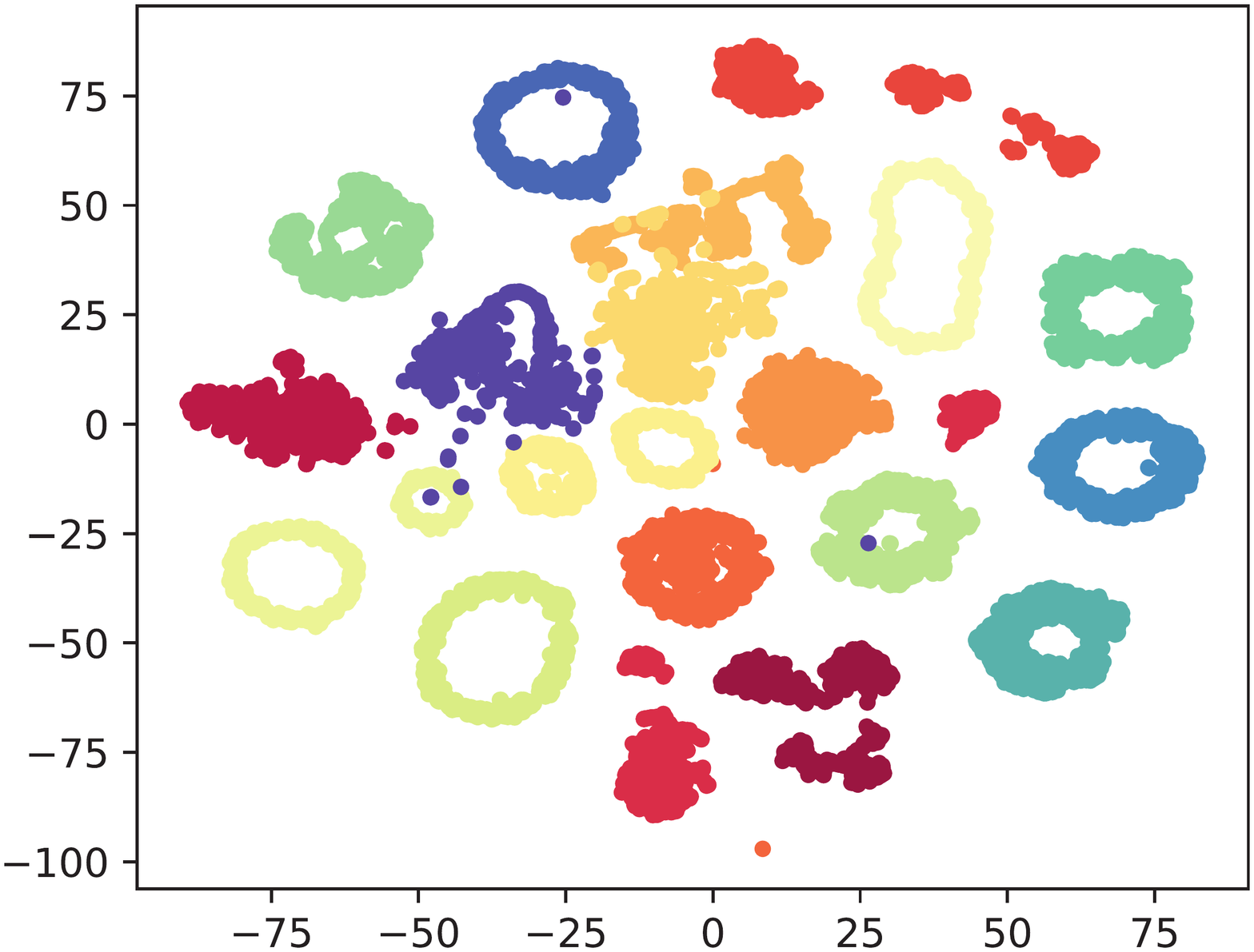}
    \centering
  \caption{UCIDSADS SAAE}
\end{subfigure}
\hspace{-1mm}
\begin{subfigure}{0.24\textwidth}
  \centering
  \includegraphics[width=\textwidth]{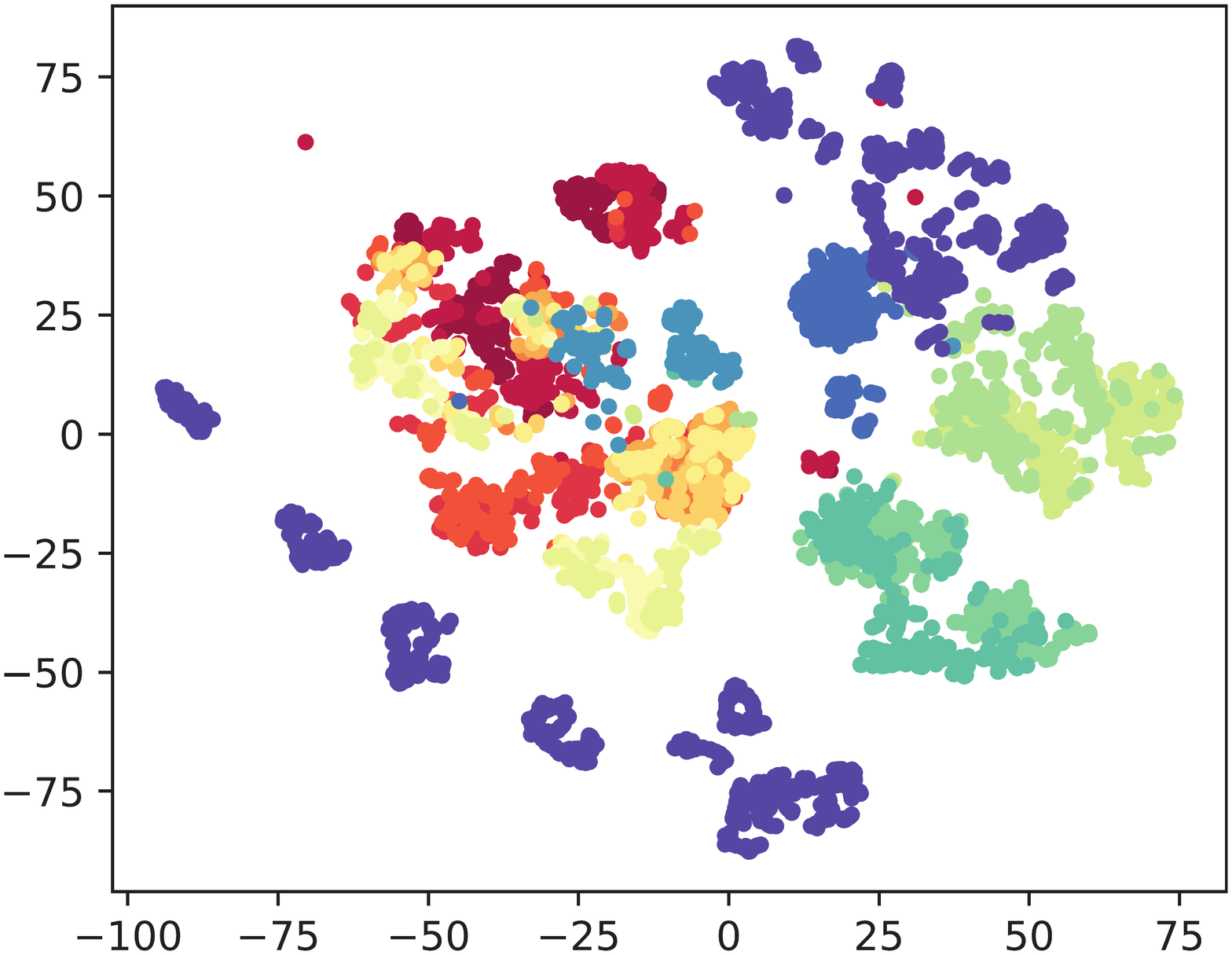}
  \caption{OPPORTUNITY Raw}
\end{subfigure}
\hspace{-1mm}
\begin{subfigure}{0.24\textwidth}
  \includegraphics[width=\linewidth]{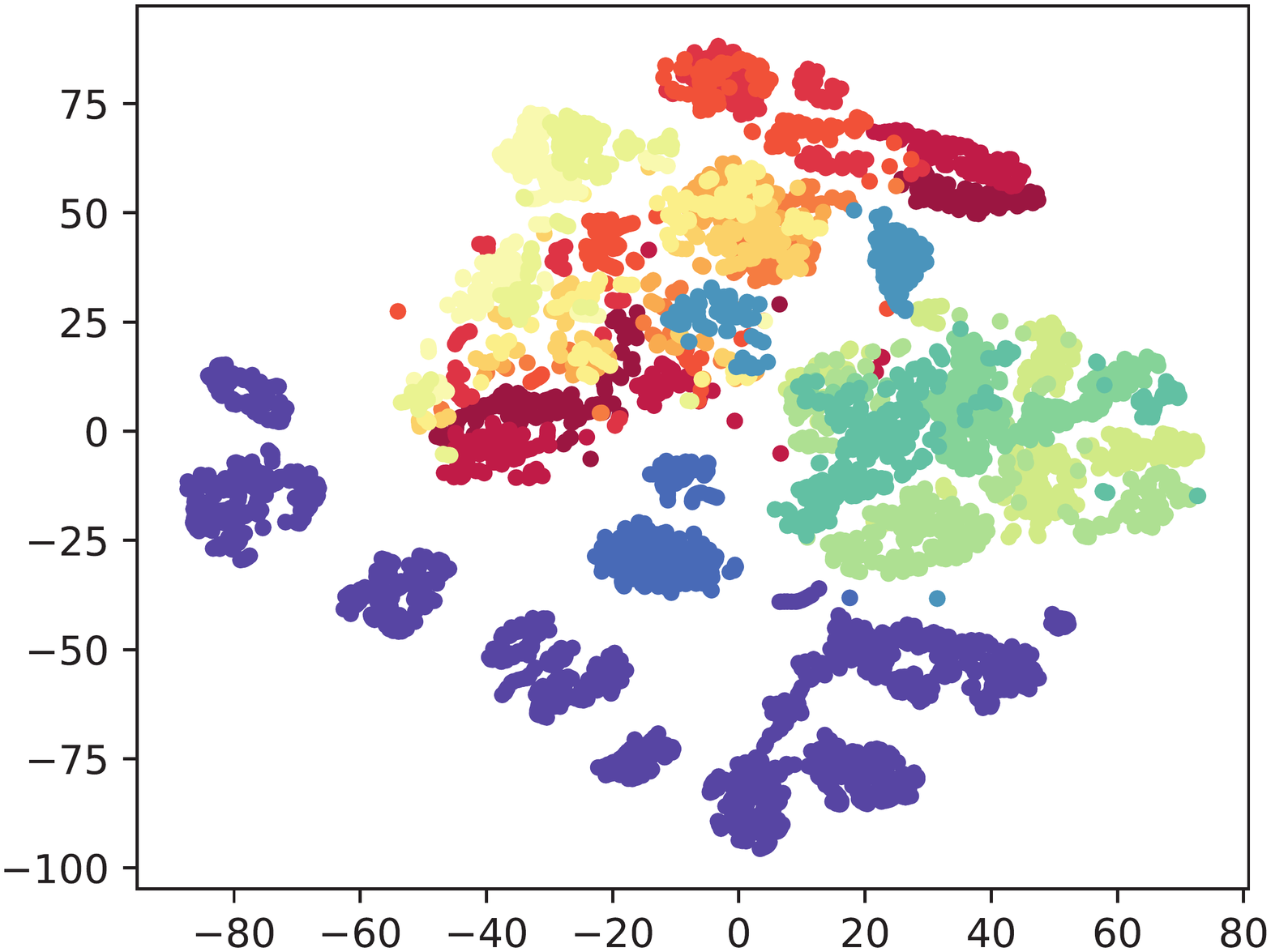}
    \centering
  \caption{OPPORTUNITY SAAE}
\end{subfigure}
\hspace{-1mm}
\caption{Performance on differnet subjects.}
\label{supplementary figure}
\end{figure*}

Due to the space limitation, we exhibit the other three datasets' analysis, i.e., specific subject accuracy reports, spectrum guided fitting curves, and embedding comparison in Fig. \ref{supplementary figure}. 

From (a)-(c), we can observe the subject variation in other three datasets, especially the subject 8 in PAMAP2 and subject 8 in UCIDSADS, while SAAE achieves stable performance over these unstable subjects. Also, SAAE improves all three subjects in OPPORTUNITY dataset. (d)-(f) further provide the discriminator fitting curves of subject 1 on other three datasets, which indicates the effectiveness of domain knowledge in diverse scenarios. (g)-(l) exhibit the data distributions before and after SAAE's purification. We can easily see that points of same classes become more gathered and there are fewer scattered points over all three datasets.

\begin{figure*}[htb]
    \centering 
\begin{subfigure}{0.23\textwidth}
  \includegraphics[width=\linewidth]{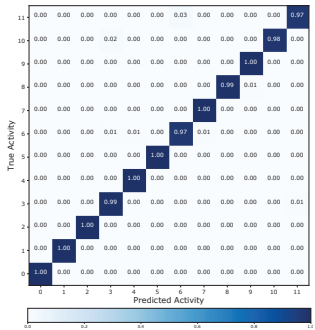}
    \centering
  \caption{MHEALTH}
\end{subfigure}\hfil 
\hspace{-1mm}
\begin{subfigure}{0.23\textwidth}
  \includegraphics[width=\linewidth]{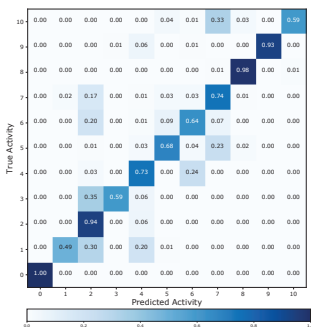}
    \centering
  \caption{PAMAP2}
\end{subfigure}\hfil 
\hspace{-1mm}
\begin{subfigure}{0.23\textwidth}
  \includegraphics[width=\linewidth]{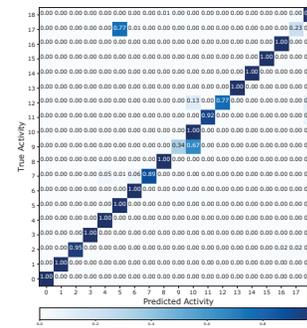}
    \centering
  \caption{UCIDSADS}
\end{subfigure}
\hspace{-1mm}
\begin{subfigure}{0.23\textwidth}
  \includegraphics[width=\linewidth,height=42mm]{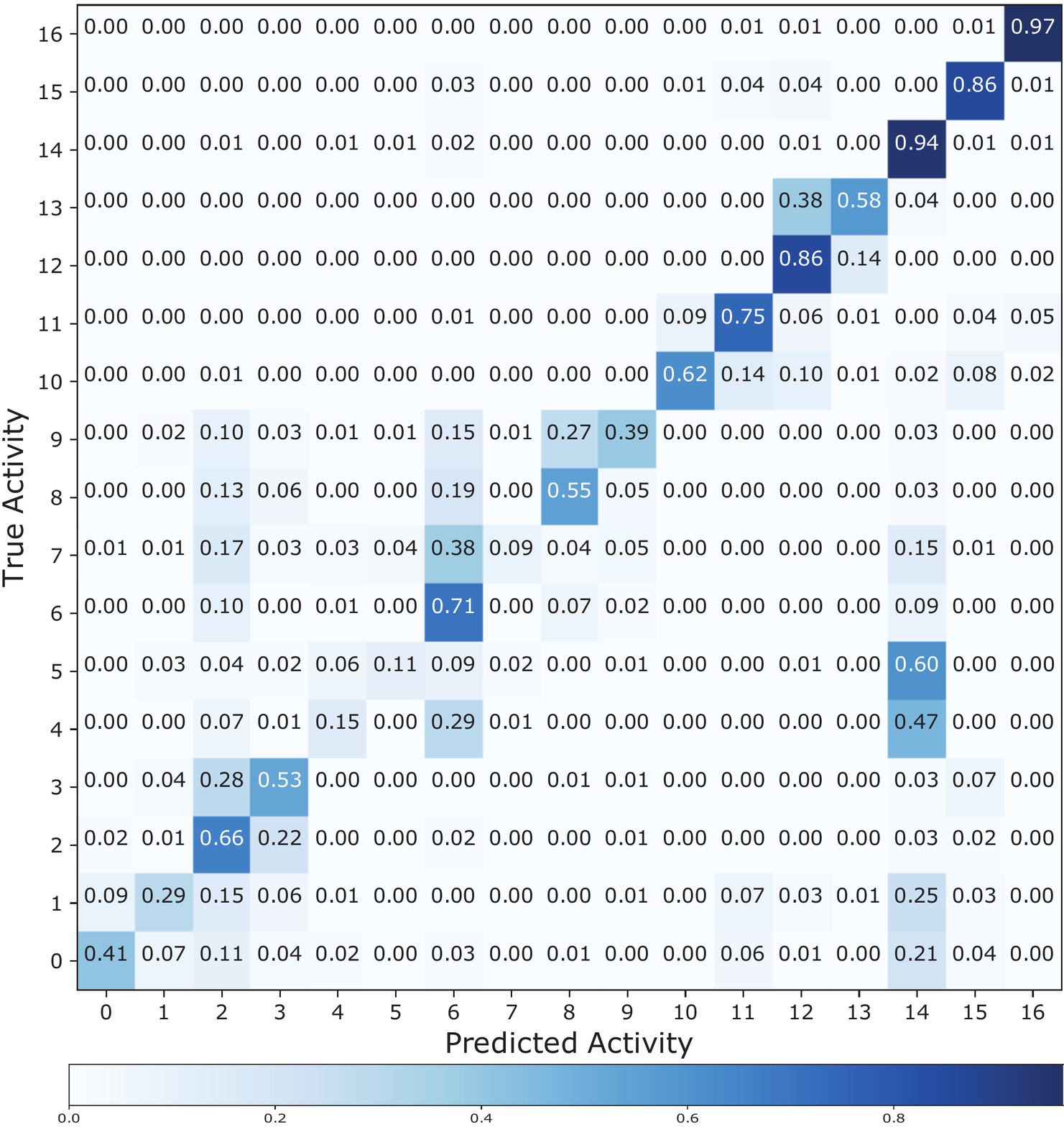}
    \centering
  \caption{OPPORTUNITY}
\end{subfigure}
\hspace{-1mm}
\caption{Confusion matrices of four datasets.}
\label{Confusion matrix}
\end{figure*}

We also exhibit the Confusion matrices over four datasets to assist proving our algorithm's outperformance and robustness.
 Fig. \ref{Confusion matrix} plots a subject's confusion matrix of the corresponding datasets. We can observe that the classification on MHEALTH is solid, and only a few samples are misclassified. The prediction in PAMAP2 shows that it is difficult to learn the generalized representations in some classes, where the misclassified samples spread over other multiple diverse activities. However, there exists the interclass similarity in UCIDSADS and OPPORTUNITY. Most of the classes can be precisely predicted in these two datasets, but the misclassified samples mainly gather in one class. For instance, the misclassified samples of class 9 (running) centers in class 10 (exercise in steps) in UCIDSADS, which indicates the interclass similarity between class 9 and class 10.

\end{document}